\documentclass{article}

\usepackage{arxiv}
\usepackage[utf8]{inputenc} 
\usepackage[T1]{fontenc}
\usepackage{xcolor}
\usepackage{adjustbox}
\usepackage{multirow}
\usepackage{hyperref}
\usepackage{cancel}
\usepackage{graphicx}
\usepackage{amsmath, amssymb, amsthm}
\usepackage{bm}
\usepackage{array}
\usepackage{caption}
\usepackage[normalem]{ulem} 
\usepackage[numbers]{natbib}

\usepackage{algorithmic}
\usepackage{algorithm}

\usepackage{subcaption}
\usepackage{soul}
\usepackage{booktabs}
\usepackage{makecell}   

\definecolor{darkgreen}{RGB}{0, 100, 0} 

\title{Physics- and geometry-aware spatio-spectral graph neural operator for time-independent and time-dependent PDEs}
\author{
  Subhankar Sarkar \\
  Yardi School of Artificial Intelligence\\
  Indian Institute of Technology Delhi, India\\
 Hauz Khas, New Delhi 110016 \\
  \texttt{subhankar.sarkar@scai.iitd.ac.in} \\
  \And
  Souvik Chakraborty \\
  Department of Applied Mechanics\\
  Yardi School of Artificial Intelligence\\
  Indian Institute of Technology Delhi\\
 Hauz Khas, New Delhi, India 110016 \\
 The Grainger College of Engineering\\ University of Illinois Urbana-Champaign,
 \\Urbana, 61801 Il, U.S.A.\\
  \texttt{souvik@am.iitd.ac.in} \\
}

\begin{document}
\maketitle




\begin{abstract}

Solving partial differential equations (PDEs) efficiently and accurately remains a cornerstone challenge in science and engineering, especially for problems involving complex geometries and limited labeled data. We introduce a Physics- and Geometry- Aware Spatio-Spectral Graph Neural Operator ($\pi$G-Sp$^2$GNO) for learning the solution operators of time-independent and time-dependent PDEs. The proposed approach first improves upon the recently developed Sp$^2$GNO by enabling geometry awareness and subsequently exploits the governing physics to learn the underlying solution operator in a simulation-free setup. While the spatio-spectral structure present in the proposed architecture allows multiscale learning, two separate strategies for enabling  geometry awareness is introduced in this paper. For time dependent problems, we also introduce a novel hybrid physics informed loss function that combines higher-order time-marching scheme with upscaled theory inspired stochastic projection scheme. This allows accurate integration of the physics-information into the loss function. The performance of the proposed approach is illustrated on number of benchmark examples involving regular and complex domains, variation in geometry during inference, and time-independent and time-dependent problems. The results obtained illustrate the efficacy of the proposed approach as compared to the state-of-the-art physics-informed neural operator algorithms in the literature.

\end{abstract}


\section{Introduction}
\label{sec:intro}
Partial differential equations (PDEs)
\cite{evans2022partial} are foundational tools across natural sciences, engineering, and technology, used to model and analyze a wide range of physical systems \cite{leung2009nonlinear}. However, analytical solutions to PDEs are rarely available, and in such cases, traditional numerical methods, such as the finite difference method (FDM) \cite{finite_diffrence}, the finite element method (FEM) \cite{finite_element}, and the finite volume method (FVM) \cite{finite_volume} are commonly employed to obtain approximate numerical solutions. These methods, while accurate, become computationally prohibitive in real-world scenarios like design optimization, where the PDE must be solved repeatedly across complex geometries and a wide range of parameter values. 
To mitigate this computational burden, scientific machine learning (SciML)-based surrogate models have been introduced \cite{sciml_1, sciml_2, sciml_3, sciml_4, surrogate4}. These models use deep neural networks to learn the mapping from inputs to outputs of a PDE, and they can generally be categorized into two classes: data-driven and physics-informed. Data-driven approaches rely on supervised learning to train deep neural networks using labeled solution data \cite{datadriven_1, datadriven_2, datadriven_3, datadriven_4, stephany2024pde, datadriven_6, datadriven_7}, but their performance heavily depends on the availability of large labeled datasets, which are often expensive or impractical to obtain. To address this issue, Physics-Informed Neural Networks (PINNs) \cite{PINN_1, PINN_2, PINN_3, PINN_4, PINN_5} have been proposed, incorporating the governing physical laws directly into the loss function, thereby eliminating the need for labeled data. While both data-driven and physics-informed approaches have shown promise, they are inherently discretization-dependent and typically require retraining for each new set of PDE parameters or changes in grid resolution, limiting their flexibility and scalability. 
Building upon the limitations of traditional neural networks for PDE surrogate modeling, neural operators \cite{FNO, SNO, CNO, OFormer, Deeponet,LNO, CNO} have emerged as a paradigm shift, aiming to learn mappings between infinite-dimensional function spaces rather than finite-dimensional input–output pairs.
Once trained, neural operators can predict solutions for previously unseen parameter configurations in a single forward pass, eliminating the need for retraining and enabling efficient, scalable PDE-solving capabilities for complex and parameterized systems.

One of the foundational works in this area is the Deep Operator Network (DeepONet) \cite{Deeponet}, inspired by the universal approximation theorem for nonlinear operators \cite{ChenandChen}. DeepONet comprises two subnetworks: a branch network that processes function values at predefined sensor locations and a trunk network that encodes the query points where the PDE solution is to be evaluated. The network outputs the solution via a dot product between the outputs of the two networks. Thanks to its operator-based formulation, DeepONet can handle parametric PDEs and irregular geometries. However, its key limitation lies in the assumption of fixed problem domain, which prevents it from generalizing to problems with varying domain geometries. PI-GANO \cite{PI-GANO} addresses this challenge by integrating a geometry encoder into the DeepONet architecture, enabling generalization across variable geometries. Despite these advances, DeepONet-based models still suffer from poor scalability with respect to the number of evaluation points, making them computationally expensive for high-resolution domains.

To overcome such scalability and generalization challenges, a variety of alternative neural operator architectures have been developed. The Graph Kernel Neural Operator (GNO) \cite{GNO} takes a graph-based approach by leveraging kernelized message passing within a graph neural network (GNN) \cite{GCN, GAT, GIN} framework. By learning a unique kernel at each graph node, GNO effectively models complex solution mappings on unstructured meshes. However, this approach incurs high memory costs, especially for large graphs, due to the need to store per-node kernels. Additionally, GNO, like other spatial GNNs, suffers from over-smoothing \cite{oversmoothing} in deeper architectures, which limits its expressivity. To address memory and scalability limitations, the Fourier Neural Operator (FNO) \cite{FNO} uses spectral convolution via the Fast Fourier Transform (FFT) to learn global kernels in frequency space. While FFT reduces spatial complexity, FNO struggles with representing localized, high-frequency phenomena due to its reliance on sinusoidal bases, often exhibiting Gibbs phenomena near discontinuities. The Wavelet Neural Operator (WNO) \cite{WNO} alleviates this by adopting multi-resolution wavelet transforms, which capture both local and global features in the solution space. Although FNO and WNO offer strong generalization and scalability on structured, regular grids, they are inherently inapplicable to irregular geometries and unstructured meshes, thereby limiting their utility in real-world applications that demand geometric flexibility.

To address the above mentioned limitations, the Spatio-Spectral Graph Neural Operator (Sp$^2$GNO) \cite{sp2gno} was introduced. Sp$^2$GNO represents the computational domain as a graph and leverages the complementary strengths of spatial and spectral graph neural networks to learn solution operators for PDEs. Spatial GNNs rely on message passing to capture short-range dependencies within a node’s local neighborhood, while spectral GNNs perform learning in the frequency domain using the Graph Fourier Transform (GFT) \cite{GFT_CMU}, allowing the model to encode long-range dependencies through the graph Laplacian's eigenbasis. By fusing these two perspectives, Sp$^2$GNO captures both fine-grained local features and global structural information, enabling it to generalize across regular and irregular domains in a seamless manner. However, like most data-driven models, Sp$^2$GNO is heavily dependent on large quantities of labeled training data, typically generated using computationally expensive numerical solvers -- ironically the very methods it aims to replace. Additionally, Sp$^2$GNO does not generalize to new geometries.

In response to these challenges, we propose a novel approach referred to as the physics- and geometry- aware spatio-spectral graph neural operator ($\pi$G-Sp$^2$GNO) for time-independent and time-dependent problems. 
$\pi$G-Sp$^2$GNO improves upon the previously proposed Sp$^2$GNO by enabling geometry awareness into the architecture while retaining its capabilities to seamlessly  (a) handle irregular domain and (b) handle unstructured grid. Two strategies for enabling geometry awareness are proposed in this paper. The first approach enables geometry awareness based on simple projection  while the second strategy employs a trainable encoder to enable geometry awareness. We also employ a novel upscaled theory based approach to encode the governing physics into the loss-function. This relaxes the necessitate associated with infinitely differentiable activation function in physics-informed networks. For time-dependent problem, we propose a hybrid strategy that combines time marching scheme with the upscaled theory for enabling physics-awareness during the training process. In a nutshell, the key features of the proposed approach are as follows:
\begin{itemize}
    \item \textbf{Geometry awareness:} The proposed $\pi$G-Sp$^2$GNO is geometry aware and hence, can zero-shot generalize to new geometries. This is a key feature that has potentially far reaching consequence in computational mechanics. 

    \item \textbf{Physics-informed learning:} $\pi$G-Sp$^2$GNO learns directly from the physics of the problems, {eliminating} the computational bottleneck associated with data generation. The hybrid loss function used for time-dependent problems also yields better convergence.

    \item \textbf{ Stochastic projection-based gradients:}  By leveraging the stochastic projection method for spatial gradient computation, $\pi$G-Sp$^2$GNO eliminates the reliance on automatic differentiation for gradient computation. This removes constraints on activation functions within the network architecture, enhancing both efficiency and flexibility.

    \item \textbf{ Diverse applicability:} $\pi$G-Sp$^2$GNO demonstrates the ability to solve a wide range of PDE problems from time-independent to time-dependent, on regular, irregular, and variable domain geometries having structured and unstructured mesh.  This broad applicability not only surpasses existing neural operators in flexibility but also extends its usability to a diverse array of real-world scientific and engineering applications.
    
\end{itemize}

The remainder of this paper is organized as follows:  The problem statement and the background is explained in Section \ref{sec:Background}. The proposed framework is described in Section \ref{sec:pi-sp2gno}. Section \ref{sec:Experiments} presents numerical experiments through which the efficiency and applicability of $\pi$G-Sp$^2$GNO are validated.
Finally, Section \ref{sec:concl} concludes the paper with final thoughts and potential future directions. 


\section{Problem Setup}
\label{sec:Background}
We consider the task of learning solution operators for parameterized partial differential equations (PDEs) defined over varying geometries using physics-informed neural operators. Let $\mathcal{D}(\mu) \subset \mathbb{R}^d$ denote a spatial domain that depends on a geometric parameter $\mu \in \mathcal{P}$, where $\mathcal{P}$ is a prescribed set of admissible shape or topology parameters. The boundary of the domain is denoted by $\partial \mathcal{D}(\mu)$, and we consider a temporal domain $\mathcal{T} \subset \mathbb{R}$ (if applicable).
The governing PDE is given in its strong form as
\begin{equation}
\label{eq:governing_pde}
    \mathcal{R}(u(\bm x, t); a(\bm x), \mu) = 0, \quad \text{for } \bm x \in \mathcal{D}(\mu), \; t \in \mathcal{T},
\end{equation}
subject to boundary and initial conditions:
\begin{align}
    \mathcal{B}(u(\bm x, t); \mu) &= 0, \quad \text{for } \bm x \in \partial\mathcal{D}(\mu), \; t \in \mathcal{T}, \\
    u(\bm x, 0) &= u_0(\bm x; \mu), \quad \text{for } \bm x \in \mathcal{D}(\mu).
\end{align}
Here, $u: \mathcal{D}(\mu) \times \mathcal{T} \to \mathbb{R}^{d_u}$ denotes the unknown solution field(s), $a: \mathcal{D}(\mu) \to \mathbb{R}^{d_a}$ represents input parameters such as material coefficients, forcing terms, or boundary data, and $\mu$ governs the geometry of the domain. The differential operator $\mathcal{F}$ encodes the physical laws and may involve both spatial and temporal derivatives.
Our goal is to learn a generalized solution operator
\begin{equation}
    \mathcal{F}: (a(\bm x), \mu) \mapsto u(\bm x, t),
\end{equation}
by a deep neural network $\mathcal{F}_\theta$ which maps the input field $a(\bm x)$ and geometric parameter $\mu$ to the corresponding PDE solution $u(\bm x, t)$ on the domain $\mathcal{D}(\mu)$.
The objective of this work is to develop a novel geometry- and physics- aware neural operator architectures capable of learning the family of solution operators $\mathcal{F}$ across a distribution of both input fields $a(\bm x)$ and geometries $\mathcal{D}(\mu)$, thereby enabling robust generalization to unseen configurations in both physics and geometry.

\section{Physics- and Geometry- Aware Spatio-Spectral Graph Neural Operator ($\pi$G-Sp$^2$GNO)}
\label{sec:pi-sp2gno}
In this section, we present our novel architecture
Physics- and Geometry- Aware Spatio-Spectral Graph Neural Operator ($\pi$G-Sp$^2$GNO)
capable of learning the solution directly from the governing laws of physics, while preserving all the properties of Sp$^2$GNO \cite{sp2gno}. To calculate the PDE loss, we exploit a novel upscaled theory based stochastic projection approach for calculating the spatial derivatives in a seamless manner. This eliminates the need for automatic differentiation, and provides flexibility regarding the choice of activation functions.

\subsection{Spatio-Spectral Graph Neural Operator (Sp$^2$GNO)}
Sp$^2$GNO \cite{sp2gno} is a kernel-based neural operator that exploits spectral and spatial graph neural networks to approximate the integral kernel. 
Spectral graph neural network captures the global long-range dependencies, while spatial graph neural network captures the short-range dependencies. Inside each Sp$^2$GNO block, we make use of these complementary properties of both GNNs by making them collaborate in such a way that the final output of each Sp$^2$GNO block is comprised of both of these complementary information in optimal proportion. Formally, The overall architecture of Sp$^2$GNO is given by,
\begin{equation}
\begin{aligned}
    v_0 (\bm x) &= P(\{a, \bm x\}) \\
    v_{t+1} (\bm x) &= W_c\left[v_{j+1}(\bm x)^{\text{spectral}} \| v_{j+1}(\bm x)^{\text{spatial}}\right], \quad \text{for $t= 1,2,..., T-1$}\\
    u (\bm x) &= Q(v_T (\bm x)),
\end{aligned}
\end{equation}
where the operations inside a graph kernel integral layer is represented as:
\begin{equation}
\begin{aligned}
    v_{t+1} (\bm x) &= W_c\left[v_{t+1}(\bm x)^{\text{spectral}} \,\|\, v_{t+1}(\bm x)^{\text{spatial}}\right], \quad \text{for } t=1,2,\dots,T-1
\end{aligned}
\end{equation}
where  $v_{j+1}(\bm x)^{\text{spectral}}$ denotes the output the spectral GNN inside $j$th Sp$^2$GNO block, and is formally expressed as:
\begin{equation}\label{eq:fast-spectral}
    v_{j+1}(\bm x)^{\text{spectral}}  =  \sigma \left(\mathbf{S_m} \cdot \mathbf{K} \times_1 (\mathbf{S_m}^\top \cdot v_j(\bm x) + w(v_j(\bm x))\right).
\end{equation}
Here $\times_1$ is the mode-1 tensor-matrix product, and $j \in \{1,2,..,L\}$ is the index for the depth of the Sp$^2$GNO block, $\sigma$ is the GELU activation function. {$\mathbf{S_m}$ is the matrix containing first $m$ eigenvectors as its columns, and $\mathbf{K} \in \mathbb{R}^{m \times d \times d}$ is the learnable 3D kernel on the spectral domain.}
On the other hand, the spatial update rule for the $j$th Sp$^2$GNO block is formulated as follows:
\begin{equation}
\label{eq:spatial_conv}
    v_{j+1}(\bm x)_u^{\text{spatial}} = \sum_{v \in \mathcal{N}(u)} \gamma_{uv} \mathbf{W} v_j(\bm x)_v
\end{equation}
where {the $\mathbf{W} \in \mathbb{R}^{d_v \times d_v}$ is the weights for the linear transformation, and} the gating coefficient $\gamma_{uv}$ is determined as:
\begin{equation}
    \gamma_{uv} =
    \begin{cases} 
          \sigma_2(\mathbf{W_3} \sigma_1 (\mathbf{W_1} [\bm{h_v}||\bm{h_u} || \mathbf{W_2}w_{uv}]))  & \text{if nodes } u \text{ and } v \text{ are connected} \\
          0 & \text{otherwise}
    \end{cases}
\end{equation}
Here, $v_{j+1}(\bm x)_u^{\text{spatial}}$ represents the updated feature vector for node $u$ in the spatial convolution operation of the $j$th Sp$^2$GNO block. The terms $\bm{h_u} \in \mathbb{R}^n$ and $\bm{h_v} \in \mathbb{R}^n$ correspond to the Lipschitz embeddings of the nodes $u$ and $v$, respectively. The neural network employs weight matrices $\mathbf{W_1}$ and $\mathbf{W_3}$, while $\mathbf{W_2}$ is the weight matrix of the linear layer. The RELU activation function is used as $\sigma_1$, and the sigmoid activation function is used as $\sigma_2$.
This convolution operation in Eq. \eqref{eq:spatial_conv} can be expressed in terms of the compact matrix notation as,
\begin{equation}
    v_{j+1}(\bm x)^{\text{spatial}} = J(v_t(\bm x))= \Gamma \odot \mathbf{A} v_j(\bm x) \mathbf{W}
\end{equation}
where $\Gamma = [\gamma_{uv}] \in \mathbb{R}^{N \times N}$ represents the learned edge weights.
This formulation integrates learned gating mechanisms into spatial graph convolutions, enhancing the model’s ability to learn the optimal graph structure by assigning high weights to important edges and nullifying the effect of unimportant edges by assigning weights close to zero. 

The vanilla  Sp$^2$GNO, as proposed in \cite{sp2gno}, is data-driven and hence, one needs to generate data using computationally expensive solvers. This can render the overall training process expensive. It is further noted that converting Sp$^2$GNO to physics-informed is not straightforward because of the presence of ReLU activation function within the architecture. 
Additionally, the aim is to make the architecture geometry aware. The strategies for integrating geometry and physics awareness into Sp$^2$GNO is discussed in subsequent sections. The resulting physics- and geometry- aware Sp$^2$GNO is referred to as the $\pi$G-Sp$^2$GNO.

\subsection{Physics Awareness}
\label{subsec:stochastic_projection}
To enable physics-awareness in Sp$^2$GNO, we proceed with the established strategy of using physics-informed loss function by using 
Eq.~\eqref{eq:governing_pde} as follows: 
\begin{equation}
\label{eq:physics_informed_loss_function}
\mathcal{L}_{\text{physics}}(\theta) = 
\underbrace{\sum_{i=1}^{N_{train}}\left\| \mathcal{R}^{(i)} \left( \bm{x}, a, t, u, \frac{\partial \mathbf{u}}{\partial t}, \frac{\partial^2 \mathbf{u}}{\partial t^2}, \dots, \frac{\partial \mathbf{u}}{\partial x}, \dots, \frac{\partial^n \mathbf{u}}{\partial x^n}  \right)\right\|_2^2}_{\text{PDE Loss} \;\mathcal{L}_{PDE}} 
+ \beta \underbrace{\sum_{i=1}^{N_{train}}\|u^{(i)} |_{\partial \mathcal{D}} - g^{(i)} \|_2^2}_{\text{Boundary Loss}}
\end{equation}
where $\beta$ is the weight of the boundary loss, $\mathbf{u}^{(i)} |_{\partial \mathcal{D}}$ predictions at the boundary vertices, and the superscript $(i)$ in $\mathcal{R}^{(i)}$ emphasize the fact that the PDE residual $\mathcal{R}$ is computed using $i$th training input. It should be noted that all $\boldsymbol{u}$ inside the loss function in Eq. ~\eqref{eq:physics_informed_loss_function} are the predicted solutions using Sp$^2$GNO, $\mathcal{F}_\theta$, i.e., $\boldsymbol{u} = \mathcal{F}_\theta(a(\bm x))$. 
The primary challenge associated with computing the residual resides in the necessity to compute the spatial derivatives.
This is particularly challenging in Sp$^2$GNO as ReLU activation is present within the Sp$2$GNO block. To address this challenge, we employ a novel upscaled theory based stochastic projection algorithm for computing the spatial derivatives, and hence, the physics-informed loss function.

The stochastic projection method \cite{NavneethStochastic} incorporates the effect of microscopic variables in the evolution of a macroscopic field variable $u(\bm x)$. On a microscopic scale, the observables evolve quickly on a smaller length scale and contain rich information about the system under consideration, which remains uncaptured for a macroscopic observer who records the observations in a much larger time and length scale. Without any knowledge of the microscopic phenomena, knowledge of the macroscopic function $u(\bm x)$ is incomplete. Based on this incomplete knowledge, the spatial variation of the macroscopic field $u(\bm y)$, at a microscopic point $\bm y$ inside an infinitesimally small neighborhood of the point $\bm x$ can be modeled by just adding a noise term
\begin{equation}
    u(\bm y) = u(\bm x) + \zeta,
\end{equation}
where $\zeta$ is a zero-mean Gaussian noise. The microscopic phenomena can be observed via the macroscopic observable $X_t$, the observation process of which is described by,
\begin{equation}
    X_t = h(\bm x- \bm y)dt + \sigma W_t,
\end{equation}
where $W_t$ is a Brownian motion independent of $\eta$, and $h$ is also a macroscopic function of the microscopic variable $(\bm x- \bm y)$. This function $h$ can not be resolved on a macroscopic scale beyond a certain threshold distance vector $\delta \hat{\bm x}$, which is the finest resolution the macroscopic observer can observe. Hence, we assume that $h$ returns zero if $|\bm y - \bm x| \leq  |\delta \hat{\bm x}|$ else, it returns a non-zero value. Complete knowledge of $u(\bm x)$ can be recovered by conditioning it on the past observations of $X_t$, taking account of the information from the microscopic scale via $X_t$. This is achieved by taking the conditional expectation of $u$ as, $u(\bm x) = \mathbb{E}_{\mathbb{P}}(u| \mathcal{F}_t ) = \Phi_t(u)$, where $\mathcal{F}_t = \sigma ( Y_s : 0 \leq s \leq t )$. This conditional expectation of any general macroscopic observable $\psi$ is calculated through another continuous distribution $\mathbb{Q}$ w.r.t $\mathbb{P}$ by the Kallianpur-Strivel formula \cite{Kallianpur_Striebel}, 
\begin{equation}
\Phi_t(\psi) = \frac{\mathbb{E}_{\mathbb{Q}}[\psi_t \Lambda_t \mid \mathcal{F}_t]}{\mathbb{E}_{\mathbb{Q}}[1 \Lambda_t \mid \mathcal{F}_t]},
\end{equation}
where $\Lambda_t$ is the Radon-Nicodym derivative $\frac{d\mathbb{P}}{d\mathbb{Q}}$ \cite{kallianpur2013stochastic}. The evolution of this conditional expectation $\Phi_t(u)$ follows the equation,
\begin{equation}
\label{eq:phi_evolution}
d\Phi_t(\psi) = \left[ \Phi_t(\psi \mathbf{h}^\top) - \Phi_t(\mathbf{h})^\top \Phi_t(\psi) \right]
\text{Var}(h)^{-1} 
\left( dY_t - \Phi_t(\mathbf{h}) \, dt \right),
\end{equation}
which can be written as,
\begin{equation}
    \begin{aligned}
    \label{eq:phi_substituted}
    d\Phi_t(\psi) &=  \Phi_t\left[(\psi - \bar{\psi})(\mathbf{h}-  \bar{\mathbf{h}})^\top\right]
\Phi_t\left[(\mathbf{h}-  \bar{\mathbf{h}})(\mathbf{h}- \bar{\mathbf{h}})^\top \right]^{-1} dX_t, \; \text{or},\\
    d\Phi_t(\psi) &= \mathbf{G} dX_t, \; \text{with}\; \mathbf{G} = \Phi_t\left[(\psi - \bar{\psi})(\mathbf{h}-  \bar{\mathbf{h}})^\top\right]
\left( \Phi_t\left[(\mathbf{h}-  \bar{\mathbf{h}})(\mathbf{h}- \bar{\mathbf{h}})^\top \right]\right)^{-1} .
    \end{aligned}
\end{equation}
Here $\bar{\psi} = \Phi_t(\psi)$, and $\bar{\psi} = \Phi_t(\psi)$. The integration due in Eq. \eqref{eq:phi_evolution} should be in the highest time and spatial resolution $\delta\hat{t}$, and $\delta \hat{\bm x}$ respectively, that a macroscopic observer can resolve. Thus, we have used $\Phi_t(h) = 0$ within the integrand in 
Eq.\eqref{eq:phi_evolution} to arrive at Eq. \eqref{eq:phi_substituted}, because, $h$ returns zero within the interval $\delta \hat{\bm x}$. Integrating Eq. \eqref{eq:phi_substituted} from initial time $t=t_0$ to $t= t_0+ \delta\hat{t}$, we obtain 
\begin{equation}
\Phi_t({\psi}) \approx \Phi_{t_0}(\psi) + \mathbf{G} \delta \hat{\bm x} ,
\end{equation}
where we have used the approximation $X_t - X_{t_0 }\approx \delta \hat{\bm x}$. Now, in the expression of the gradient $\mathbf{G}$, we can substitute $\psi = u(\bm x)$, $\bar{\psi} = u(\bar{\bm x})$, $h = \bm x-\bar{\bm x}$,
\begin{equation}
    \mathbf{G}(\bar{\bm x}) = \frac{\Phi_t \left[(u(\bm x) - u(\bar{\bm x}))(\bm x - \bar{\bm x})^\top\right]}{\Phi_t \left[(\bm x - \bar{\bm x})(\bm x - \bar{\bm x})^\top\right]}.
\end{equation}
The Monte Carlo approximation of this gradient is then calculated choosing $N_b$ number of points within a ball shaped neighborhood centered at every $\bar{\bm x}$,
\begin{equation}
\label{eq:stochatic_projection_gradients}
    \mathbf{\hat{G}}(\bar{\bm x}) = \frac{\partial u(\bar{\bm x}, \theta)}{\partial \bm x} = 
\frac{\frac{1}{N_b} \sum_{i=1}^{N_b} ( u( \bm{x}_i, \theta) - u(\bar{\bm x}) ) (\bm{x}_i - \bar{\bm x})^T}
{\frac{1}{N_b} \sum_{i=1}^{N_b} (\bm{x}_i - \bar{\bm x}) (\bm{x}_i - \bar{\bm x})^T}.
\end{equation}
Assuming time-independent problem, Eq. \eqref{eq:stochatic_projection_gradients} can be directly used to compute the spatial derivatives, which can then be exploited to formulate the physics-informed loss function in Eq. \eqref{eq:physics_informed_loss_function}. Addressing time-dependent problem poses additional challenge associated with the temporal derivative. This has been tackled separately in a later section.

\subsection{Geometry Awareness}\label{subsec:geom_aware}
Sp$^2$GNO, being a graph-based neural operator, inherently exhibits a degree of geometry awareness due to its reliance on graph representations. However, the presence of spectral components within the Sp$^2$GNO blocks necessitates that node features be non-zero in order to effectively encode and propagate geometric information. When the input coefficients or source terms are defined throughout the domain, including both boundary and interior nodes, the inherent geometric inductive bias of Sp$^2$GNO suffices to capture variations in domain geometry. However, in many practical physical scenarios where inputs are specified only at the boundaries, the interior node features remain zero, thereby limiting the model’s ability to learn geometric dependencies.
To overcome this limitation, we propose two complementary strategies to enrich the node feature representation and enhance geometry encoding.

\textbf{Strategy 1: Interpolation-based Enrichment}: The first strategy involves interpolating the boundary condition data onto the interior nodes, thereby producing non-zero node features across the entire graph. These interpolated features are then passed through a simple linear encoder layer \( P \), which lifts them to a higher-dimensional representation suitable for input into the subsequent $\pi$G-Sp$^2$GNO blocks. This approach ensures that geometric information is encoded at every node, enabling the operator to effectively utilize the full spatial context of the problem. We refer to this architecture as {$\pi$G-Sp$^2$GNO} (see Fig.~\ref{fig:piG_sp2gno_int_sub}).

\textbf{Strategy 2: Learnable Encoder Network}: The second strategy adopts a learnable encoder-decoder architecture designed to construct informative node features while preserving the geometric structure of the domain. To maintain the functional inductive bias characteristic of neural operators, we define an encoder network \( P \) that extracts spatially meaningful features from both the boundary condition values and the nodal coordinates.
In this framework, the spatial coordinates \( \bm{x} \in \mathbb{R}^{N \times d} \) and the boundary condition values \( u_{BC} \in \mathbb{R}^{N_{BC} \times d_u} \) are processed in parallel using two separate encoders: the coordinate encoder \( \mathcal{P}_{\text{coor}} \) and the boundary condition encoder \( \mathcal{P}_{\text{BC}} \). Their outputs, \( \tau \in \mathbb{R}^{N \times d_{\text{hidden}}} \) and \( \beta \in \mathbb{R}^{d_{\text{hidden}}} \), respectively are concatenated and passed through a nonlinear decoder \( F_1 \) (decoder-1 in Fig.~\ref{fig:piG_sp2gno_enc_sub}) to yield the encoded representation \( \mathbf{H}_1 \in \mathbb{R}^{N \times d_v} \), which serves as input to the downstream $\pi$G-Sp$^2$GNO layers:
\begin{equation}
    \begin{aligned}
    \beta &= \mathcal{P}_{BC}(u_{BC}),\\
    \tau &= \mathcal{P}_{coor}(\bm{x}),\\
    \mathbf{H}_1 := P(\bm{x}, u_{BC}) &= F_1(\tau, \beta) = F_1(\mathcal{P}_{coor}(\bm{x}), \mathcal{P}_{BC}(u_{BC})).
    \end{aligned}
\end{equation}
The decoder \( F_1 \) is implemented as a neural network with two hidden layers to capture complex interactions between the coordinate and boundary condition embeddings.

\textbf{Extension to Geometry-Dependent Problems}:
For problems where the domain geometry varies across samples, we further extend this architecture by incorporating a geometry encoder \( \mathcal{P}_{\text{geo}} \). This encoder processes the geometric context using the concatenated boundary condition values and their corresponding spatial coordinates:
\begin{equation}
    \zeta = \mathcal{P}_{geo}(\{u_{BC}, x_{BC}\}),
\end{equation}
where \( x_{BC} \in \mathbb{R}^{N_{BC} \times d} \) denotes the boundary coordinates. The resulting geometry encoding \( \zeta \in \mathbb{R}^{d_{hidden}} \) is fused with the earlier encoded output \( \mathbf{H}_1 \) via an element-wise (Hadamard) product and passed through an additional decoder \( F_2 \) (decoder-2 in Fig.~\ref{fig:piG_sp2gno_enc_sub}) to yield the final encoded representation \( \mathbf{H}_2 \in \mathbb{R}^{N \times d_v} \) for input into the subsequent $\pi$G-Sp$^2$GNO blocks:
\begin{equation}
    \mathbf{H}_2 := P^{geo}(\bm{x}, u_{BC}) = F_2(\mathbf{H}_1 \odot \zeta).
\end{equation}
We refer to this extended architecture as {$\pi$G-Sp$^2$GNO-Encoder} (Fig.~\ref{fig:piG_sp2gno_enc_sub}). Overall, the encoded output \( \mathbf{H}_1 \) is used for problems where the geometry is fixed, while \( \mathbf{H}_2 \) is used when the geometry varies across samples.

\subsection{$\pi$G-Sp$^2$GNO for time-independent systems}
\label{subsec:model_time_independent}
\begin{figure}[ht!]
    \centering
    \begin{subfigure}[b]{0.95\textwidth}
        \centering
        \includegraphics[width=\textwidth]{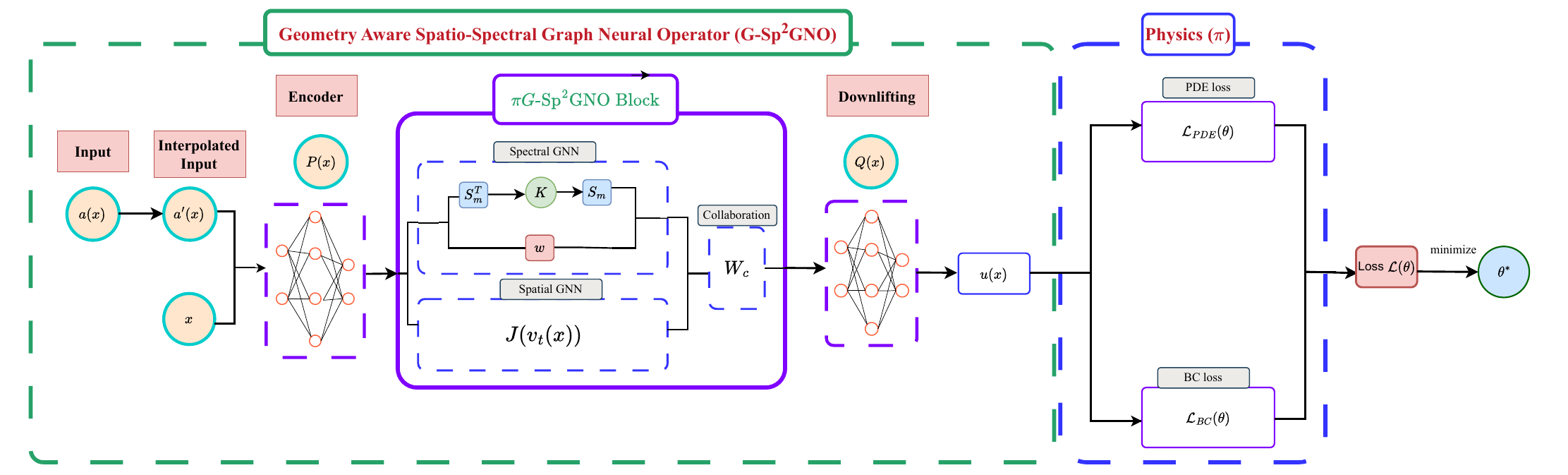}
        \caption{$\pi$G-Sp$^2$GNO framework with interpolated input construction and encoding using a linear encoder.}
        \label{fig:piG_sp2gno_int_sub}
    \end{subfigure}
    
    \vspace{1em}
    
    \begin{subfigure}[b]{0.95\textwidth}
        \centering
        \includegraphics[width=\textwidth]{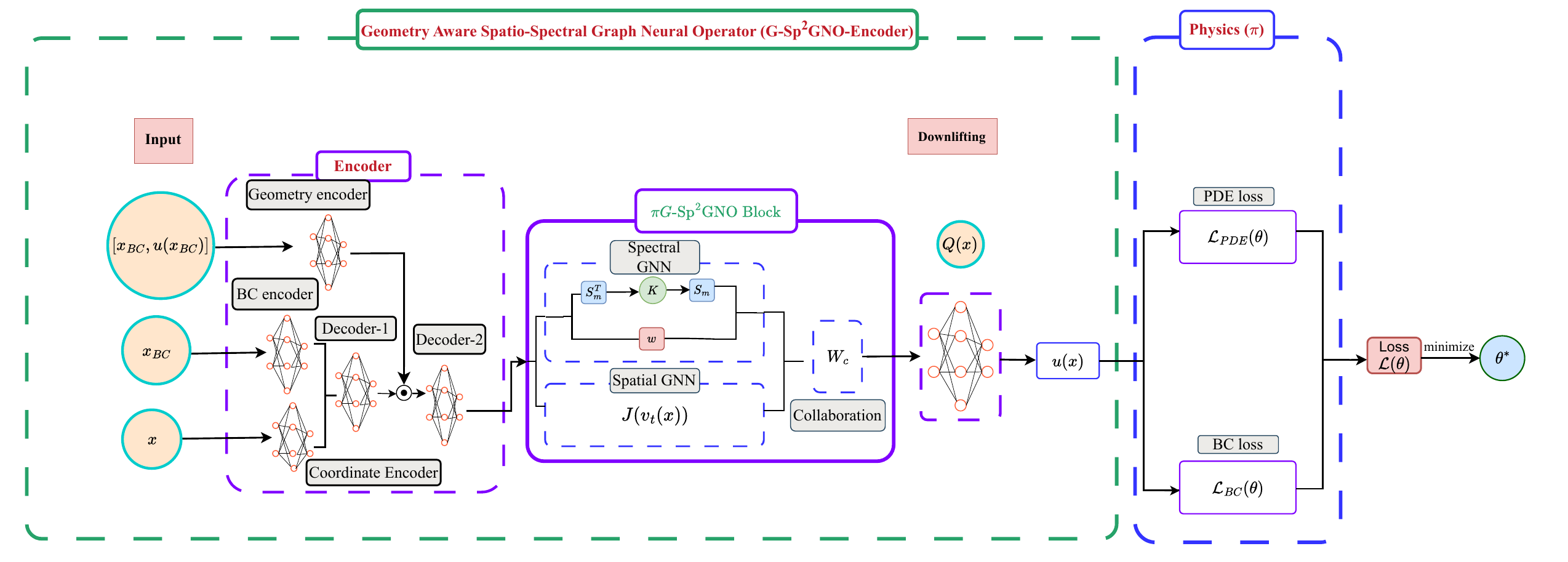}
        \caption{$\pi$G-Sp$^2$GNO-Encoder framework with geometric projection using a unified encoder module.}
        \label{fig:piG_sp2gno_enc_sub}
    \end{subfigure}
    
    \caption{Overall architecture of the proposed Physics- and Geometry-Aware Spatio-Spectral Graph Neural Operator ($\pi$G-Sp$^2$GNO) with their two variants. 
    (a) The $\pi$G-Sp$^2$GNO framework with interpolated input construction and a linear encoder for uplifting the input. 
    (b) The $\pi$G-Sp$^2$GNO-Encoder  framework with a geometry-aware encoder that projects inputs onto a unified grid. 
    Both versions share the core Sp$^2$GNO architecture, which learns the input–output mapping in spatio-spectral space, and incorporate a physics-informed training component that enables direct supervision from governing physical laws. Gradient computation within the physics-informed loss is facilitated by an upscaled theory-inspired stochastic projection algorithm.}
    \label{fig:piG_sp2gno_combined}
\end{figure}

Having discussed strategies to incorporate both physics and geometry awareness, we now turn our attention to the resulting $\pi$G-Sp$^2$GNO framework and its application to time-independent problems.
These time-independent problems can be broadly categorized into three types, based on the nature of the input provided to the model. In all cases, the output is represented as $u(\bm{x}) \in \mathbb{R}^{N \times d_u}$, defined over the entire spatial domain. However, the key distinction between these problem types lies in how the input 
\( a(\bm{x}) \) 
is defined.
In the first category, the input consists solely of boundary conditions \( g(\bm{x}) \in \mathbb{R}^{N_{BC} \times d_a} \), \( \bm{x} \in \partial D \). Each training and test instance may have different boundary conditions, but the domain remains unchanged. Since the geometry does not vary, geometry awareness is not required in this setting. It is important to note that the boundary conditions must be interpolated throughout the entire domain. For instance, consider a triangular domain with distinct boundary conditions \( g_i(\bm{x}) \) for each edge \( i = 1, 2, 3 \). In this case, the input to the model is defined as \( a(\bm{x}) \in \mathbb{R}^{N \times 3d_a} \), where at any interior point \( \bm{x}_i \), the input vector is constructed as: \( a(\bm{x}_i) = \mathbin{\|}_{j=1}^{d_a}( g_1^j(\bm{x}_i), g_2^j(\bm{x}_i), g_3^j(\bm{x}_i) ) \). Here, \( \mathbin{\|} \) denotes the concatenation operator across boundary segments, and \( g_i(\bm{x}) = (g_i^1(\bm{x}), g_i^2(\bm{x}), \dots, g_i^{d_a}(\bm{x})) \) corresponds to the \( d_a \)-dimensional boundary condition on segment \( i \).
In the second category, both the boundary conditions and the domain geometry are allowed to vary. As a result, each training and test sample includes distinct boundary conditions defined on different geometries. This setting necessitates that the model be both physics- and geometry-aware.
In the third category, the input is defined over the entire spatial domain rather than just on the boundary. Examples of such problems include those where global system parameters or spatially distributed forcing functions serve as input. For this scenario, only physics-aware Sp$^2$GNO suffices. A schematic representation of the overall framework is shown in Fig. \ref{fig:piG_sp2gno_combined}.



\begin{algorithm}[ht]
\caption{Physics- and Geometry- aware Spatio-Spectral Graph Neural Operator .}
\label{alg:piG-sp2gno_tind}
\textbf{Requirements:} Boundary conditions, initial conditions, and PDE describing the physics constraint.
\begin{algorithmic}[1]
\STATE \textbf{Initialize:} Network parameters of the Sp$^2$GNO, $\bm \theta$.
\STATE Compute the output of Sp$^2$GNO $u(\bm x)$.
\STATE Define the neighborhood of each point in the domain based on a radius $r$, and select the number of neighborhood points $N_b$ for gradient computation.
\STATE Obtain the first-order gradients at all the collocation points using Eq.~\eqref{eq:stochatic_projection_gradients} and store the gradients.
\STATE From the first-order gradients, using the same formulation in Eq.~\eqref{eq:stochatic_projection_gradients}, compute the second-order gradients.
\STATE Compute the PDE loss $\mathcal{L}_{PDE}$ defined in Eq.~\eqref{eq:physics_informed_loss_function} using the calculated gradients.
\STATE Compute the boundary loss $\mathcal{L}_{BC}$. Sum all the losses to get the total loss $\mathcal{L}_{physics}$ as shown in Eq.~\eqref{eq:physics_informed_loss_function}.
\WHILE{$\mathcal{L} > \epsilon$}
    \STATE Train the network: $ \bm{\theta} \leftarrow \bm \theta - \eta \nabla_{ \bm \theta} \mathcal{L}(\bm \theta)$
    \STATE epoch $\gets$ epoch $+ 1$
\ENDWHILE
\STATE Return the optimum parameters $\bm \theta^*$ for the network.
\STATE Obtain predictions/solutions.
\end{algorithmic}
\textbf{Output:} Prediction of the field variable/solution of PDE.
\end{algorithm}

\subsection{$\pi$G-Sp$^2$GNO for spatio-temporal systems}
\label{subsec:problem_setup_for_time_depndent}

Let us consider the following generalized time-dependent partial differential equation (PDE):
\begin{equation}
\label{eq:time_dependent_problem}
    \mathcal{R} \left( u, a, \bm x, t, \frac{\partial u}{\partial \bm x}, \frac{\partial^2 u}{\partial \bm x^2},\cdots,\frac{\partial^n u}{\partial \bm x^n} \right) := \frac{\partial u}{\partial t} - F \left( u, a, \bm x, t, \frac{\partial u}{\partial \bm x}, \frac{\partial^2 u}{\partial \bm x^2},\cdots,\frac{\partial^n u}{\partial \bm x^n} \right) = 0.
\end{equation}
A fundamental challenge associated with time-dependent systems lies in handling the temporal derivative. Direct application of automatic differentiation or stochastic projection to compute this term can be nontrivial. To address this, we propose a hybrid approach that integrates a time-marching scheme with the stochastic projection framework introduced earlier. The key idea is to utilize stochastic projection for efficiently computing spatial derivatives while employing a time-marching strategy to evolve the system temporally.
This formulation enables the proposed \(\pi\text{G-Sp}^2\text{GNO}\) model to learn the dynamics of the system and predict future states in an auto-regressive fashion, starting from an initial condition. To ensure a fixed input sequence length of \( k \), \(\pi\text{G-Sp}^2\text{GNO}\) augments its inputs with the initial condition \( u_0 \), repeated \( k - n \) times whenever fewer than \( k \) past predictions are available (i.e., for \( n < k \)).
Solving the time-dependent PDE involves evolving the system from time \( t \) to \( t + \Delta t \), analogous to traditional numerical solvers. \(\pi\text{G-Sp}^2\text{GNO}\) is trained to perform this single-step integration using the governing physics embedded in the PDE loss. Once trained, the model learns the underlying system dynamics and can recursively predict the response \( [u_1, u_2, \dots, u_N] \) using only the initial state \( u_0 \).
This is achieved by learning the solution operator \( \mathcal{F} : [u_n, u_{n-1}, \dots, u_{n-k}] \mapsto u_{n+1} \), which maps the previous \( k \) predictions to the next state. The corresponding formulation is:
\begin{equation}
\label{eq:problem_formulation}
u_{n+1} = \mathcal{F}_{\bm \theta}(\bm{a}_{n+1}), \quad \bm{a}_{n+1} = 
\begin{cases}
\left[ u_n, u_{n-1}, \dots, u_{n-k} \right] \,\Vert\, 
[ \underbrace{u_0, \dots, u_0}_{k - n\ \text{times}} ], & \text{if } n \leq k, \\[10pt]
\left[ u_n, u_{n-1}, \dots, u_{n-k} \right], & \text{if } n > k.
\end{cases}
\end{equation}
As \( n \) increases, the model progressively updates the input sequence by incorporating the most recent prediction \( u_n \). For example, the prediction \( u_1 \) is computed using the input sequence \( [u_0, u_0, \dots, u_0] \) repeated \( k \) times. After obtaining \( u_1 \), it is appended to the sequence to predict the subsequent step. This process continues recursively up to \( N \) time steps:
\begin{equation}
\label{eq:recursive_solving}
\begin{aligned}
u^1 &= \mathcal{F}_{\bm \theta}\left( \bm{a}_1\right), \quad && \bm{a}_1 = [ u_0, u_0, u_0, \dots, u_0 ], \\
u^2 &= \mathcal{F}_{\bm \theta} \left( \bm{a}_2\right), \quad && \bm{a}_2 = \left[ u_1, u_0, u_0, \dots, u_0 \right], \\
u^3 &= \mathcal{F}_{\bm \theta} \left( \bm{a}_3\right), \quad && \bm{a}_3 = \left[ u_2, u_1, u_0, \dots, u_0 \right], \\
&\vdots \\
u_N &= \mathcal{F}_{\bm \theta} \left( \bm{a}_N \right), \quad && \bm{a}_N = \left[ u_{N-1}, u_{N-2}, \dots, u_{N-1-k} \right].
\end{aligned}
\end{equation}
Thus, the prediction at time \( n+1 \) can be expressed as a function of the past \( k \) predictions, which in turn are recursively derived from the initial condition \( u_0 \):
\begin{equation}
\label{eq:overall_formulation}
u_{n+1} = \hat{\mathcal{F}_{\bm \theta}}(u_0) = \mathcal{F}_{\bm \theta} \left( \left[ 
\mathcal{F}_{\bm \theta}(\bm{a}_n),\ 
\mathcal{F}_{\bm \theta}(\bm{a}_{n-1}),\ 
\dots,\ 
\mathcal{F}_{\bm \theta}(\bm{a}_{n-k}) 
\right] \right).
\end{equation}

Although a variety of time-marching schemes can be employed, we utilize the Crank--Nicolson method \cite{Crank_Nicolson_1947}:
\begin{equation}
\label{eq:crank_nicolson}
\frac{u^{n+1} - u^n}{\Delta t} 
= \frac{1}{2} \left[
F^{n+1} \left( u, \bm x, t, \frac{\partial u}{\partial \bm x}, \frac{\partial^2 u}{\partial \bm x^2} \right)
+ F^n \left( u, \bm x, t, \frac{\partial u}{\partial \bm x}, \frac{\partial^2 u}{\partial \bm x^2} \right)
\right].
\end{equation}
The corresponding PDE loss is then defined as:
\begin{equation}
\label{eq:time_dependent_pde_loss}
\mathcal{L}_{PDE}=\sum_{i=1}^B\sum_{n=1}^{N-1}\left\|
u_i^{n+1} - u_i^n 
- \frac{\Delta t}{2} \left[
F_i^{n+1} \left( u, \bm x, t, \frac{\partial u}{\partial \bm x}, \frac{\partial^2 u}{\partial \bm x^2} \right)
+ F_i^n \left( u, \bm x, t, \frac{\partial u}{\partial \bm x}, \frac{\partial^2 u}{\partial \bm x^2} \right)
\right]
\right\|^2.
\end{equation}
Here, the superscripts \( n \) and \( n+1 \) in \( F^n \) and \( F^{n+1} \) indicate that the function \( F \) is evaluated at time steps \( n \) and \( n+1 \), respectively. The subscript \( i \) denotes the index of the training samples in a batch of size \( B \). The spatial derivatives \( \frac{\partial u}{\partial \bm x} \) and \( \frac{\partial^2 u}{\partial \bm x^2} \) are computed using the stochastic projection approach discussed in subsection \ref{subsec:stochastic_projection}.

An outline of the complete procedure for solving time-dependent PDEs using \(\pi\text{G-Sp}^2\text{GNO}\) is presented in Algorithm~\ref{alg:sp2gno_td}.

\begin{algorithm}[H]
\caption{Training $\pi$G-Sp$^2$GNO for Time-Dependent PDEs}
\label{alg:sp2gno_td}
\begin{algorithmic}[1]
\REQUIRE Initial conditions $\{u_0^{(i)}\}_{i=1}^{B}$, sequence length $k$, number of time steps $N$, time step $\Delta t$, number of training epochs $E$
\STATE \textbf{Initialize:} Network parameters of the Sp$^2$GNO, $\bm \theta$.

\FOR{epoch = 1 to $E$}
    \STATE Initialize total loss: $\mathcal{L}_{\text{PDE}} \gets 0$

    \FOR{each training sample $i = 1$ to $B$}
        \STATE Set $u_0^{(i)}$ as initial state
        \STATE Initialize predicted sequence: $\bm{U}^{(i)} \gets [u_0^{(i)}]$

        \FOR{$n = 0$ to $N-1$}
            \IF{$n < k$}
                \STATE Construct input $\bm{a}_{n+1}^{(i)} = [u_n^{(i)}, u_{n-1}^{(i)}, \ldots, u_0^{(i)}, u_0^{(i)}, \ldots, u_0^{(i)}]$ \COMMENT{Pad to length $k$}
            \ELSE
                \STATE Construct input $\bm{a}_{n+1}^{(i)} = [u_n^{(i)}, u_{n-1}^{(i)}, \ldots, u_{n-k}^{(i)}]$
            \ENDIF

            \STATE Predict next state: $u_{n+1}^{(i)} = \mathcal{F}_{\bm \theta}(\bm{a}_{n+1}^{(i)})$
            \STATE Append $u_{n+1}^{(i)}$ to $\bm{U}^{(i)}$

            \STATE Compute spatial derivatives $\partial_x u$, $\partial_{xx} u$ using stochastic projection

            \STATE Compute PDE residual: $ r_{n+1}^{(i)} = u_{n+1}^{(i)} - u_n^{(i)} - \frac{\Delta t}{2} \left[ F^{n+1}_i + F^n_i \right]$
             \STATE Compute BC loss: $\mathcal{L}_{BC}= \|u^{(i)} |_{\partial \mathcal{D}} - g^{(i)} \|_2^2$
            \STATE Accumulate PDE loss: $\mathcal{L}_{\text{PDE}} \mathrel{+}= \|r_{n+1}^{(i)}\|^2$
            \STATE Accumulate BC loss: $\mathcal{L}_{\text{BC}} \mathrel{+}= \|\mathcal{L}_{BC}\|^2$
        \ENDFOR
    \ENDFOR
    \STATE calculate physics informed loss : $ \mathcal{L}_{physics} = \mathcal{L}_{PDE}+ \beta \mathcal{L}_{BC}$
    \STATE Update model parameters $\bm \theta$ using gradient descent on $\mathcal{L}_{\text{physics}}$ as: $ \mathbf{\bm \theta} \leftarrow \bm \theta - \eta \nabla_{\bm \theta} \mathcal{L}_{physics}(\bm \theta)$
\ENDFOR

\RETURN Trained model $\mathcal{F}_{\bm \theta}$
\end{algorithmic}
\end{algorithm}

\section{Numerical experiments}
\label{sec:Experiments}
In this section, we present a comprehensive numerical evaluation of the proposed architecture with respect to accuracy, generalizability, and computational efficiency across a suite of benchmark PDE problems that typify complex physical and engineering systems. To systematically demonstrate the robustness and versatility of our approach, we consider eight representative examples organized into three categories: (a) problems defined on fixed domain geometries, (b) problems involving variable domain geometries, and (c) time-dependent problems.
Our evaluation combines both qualitative and quantitative analyses. Spatial fidelity of the predicted solutions is assessed through contour plots, providing intuitive visualization of the model's performance, while relative mean squared error (MSE) values, reported in Table~\ref{table:performance_comparison}, quantify predictive accuracy. To ensure a rigorous and equitable comparison, we benchmark the proposed method against several state-of-the-art physics-informed neural operator models namely, PI-DCON~\cite{PI-DCON}, PI-GANO~\cite{PI-GANO}, and PI-DeepONet~\cite{PI-DeepONet}, which are specifically designed to handle complex geometries and unstructured computational domains.

For all test cases, the spatial domains are discretized as unstructured point clouds, and the corresponding graph structures are constructed using the \textit{k}-nearest neighbors  algorithm, with the number of neighbors \( k \) chosen based on problem-specific considerations. All nonlinear transformations utilize the Gaussian Error Linear Unit activation function. Model training is performed using the ADAM optimizer, with an initial learning rate of \( 0.001 \), which is decayed by a factor of \( 0.65 \) every 100 epochs. Additional problem-specific hyperparameters and implementation details are outlined in the respective subsections.

\subsection{Example Set I: Time-independent problems defined on fixed geometry}
As the first set of examples, we consider problems defined on fixed geometries, where the objective is to learn the mapping from spatially varying input parameters to the corresponding response fields. This category includes three representative examples. We evaluate the predictive performance of the proposed \(\pi\)G-Sp\(^2\)GNO architecture and compare it against several state-of-the-art approaches reported in the literature.
\subsubsection{Non-homogeneus Poisson's Equation}

As the first example, we consider the 2D Poisson equation with periodic boundary conditions \cite{PIWNO},
\begin{equation}
\label{eq:poissons}
    \begin{aligned}
   \frac{\partial^2 u}{\partial  x^2} + \frac{\partial^2 u}{\partial y^2} &= f( x, y), \quad x, y \in [-1,1]^2 \\
   u( x = -1, y) &= u( x = 1, y) = u( x, y = -1) = u(x, y = 1) = 0,
    \end{aligned}
\end{equation}
where $u$ is the field variable and $f(x, y)$ is the source term. This is a time-independent problem where input source terms vary across the training and test examples while the domain geometry and boundary conditions remain constant across all examples. 
Considering the analytical solution, 
\begin{equation}\label{eq:poissons_u}
    u( x,  y) = \alpha \sin(\pi x)(1 + \cos(\pi y)) + \beta \sin(2\pi x)(1 - \cos(2\pi y)),
\end{equation}
the source term can be represented as,
\begin{equation}\label{eq:poissons_f}
\begin{split}
    f( x,  y) = & 16\beta\pi^2 \left( \cos(4\pi y) \sin(4\pi  x) + \sin(4\pi x)(\cos(4\pi y) - 1) \right) - \\
    & \alpha\pi^2 \left( \cos(\pi y) \sin(\pi x) + 
 \sin(\pi x)(\cos(\pi y) + 1) \right).
\end{split}
\end{equation}
\begin{figure}[ht]
  \centering
  \includegraphics[width=1.0\textwidth]{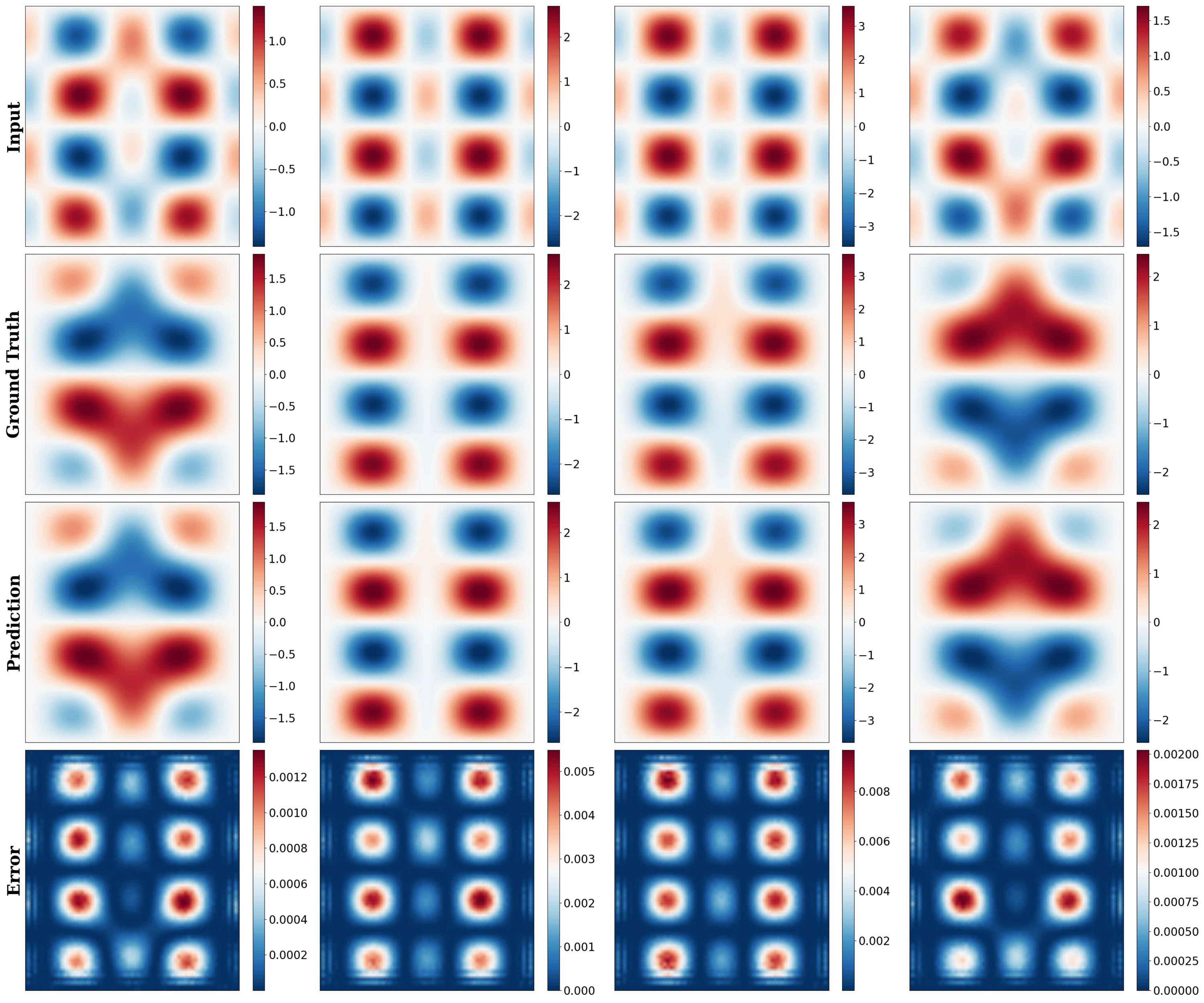}
  \caption{\textbf{Predictions for non-homogeneous Poisson's equation}: The input source function, ground truth, predictions, and errors of four different test examples are plotted in four columns. The contour plots in the first row represent the four examples' source terms. The respective ground truth and predictions for the same are plotted in the 2nd and 3rd rows, respectively. The corresponding point-wise squared errors are plotted in the fourth row.}
  \label{fig:poissons}
\end{figure}
The objective here is to learn a neural operator $\mathcal{F}_{\bm \theta}: f( x,y) \mapsto u(x, y)$ directly from the governing physics defined in Eq. \eqref{eq:poissons}. $f( x,y)$ required for training the model are generated by sampling $\alpha \sim \text{Unif}(-2,2)$, and $\beta \sim \text{Unif}(-2,2)$, and substituting the same in Eq. \eqref{eq:poissons_f}. We note that as the proposed model is directly trained from the governing physics, it does not require samples for $u(x,y)$ during the training process.

Fig. \ref{fig:poissons} (third row) shows the response obtained using the proposed $\pi$G-Sp$^2$GNO corresponding to four randomly generated $f( x,y)$. The results obtained are compared against ground truth results obtained using Eq. \eqref{eq:poissons_u} (second row of Fig. \ref{fig:poissons}). 
Qualitatively (visually) and quantitatively (fourth row of Fig. \ref{fig:poissons}), it is evident that the proposed approach yields highly accurate results matching almost exactly with the ground truth results. 
Table \ref{table:performance_comparison} illustrate the performance of the proposed $\pi$G-Sp$^2$GNO architecture by comparing its Normalized Mean Squared Error (N-MSE) with other baseline models. 
The results demonstrate that the N-MSE error of the proposed approach is significantly smaller when compared to the other baselines. The marked improvement in performance with $\pi$G-Sp$^2$GNO suggests that the architecture not only learns the solution operator more effectively but also exhibits better generalization and accuracy in predicting the solution fields.

\begin{table}[ht]
  \caption{Performance comparison with different baselines on selected benchmarks. N-MSE is reported. A smaller N-MSE indicates better performance. Two variants of our framework, $\pi$G-Sp$^2$GNO and $\pi$G-Sp$^2$GNO-Encoder are shown. 
   } 
  \label{table:performance_comparison}
  \centering
  \begin{adjustbox}{max width=\textwidth}
    \begin{tabular}{
      @{\hskip 2pt}l@{\hskip 8pt} 
      c@{\hskip 3pt}c@{\hskip 3pt}c@{\hskip 12pt} 
      c@{\hskip 3pt}c@{\hskip 12pt} 
      c@{\hskip 3pt}c@{\hskip 3pt}c@{\hskip 3pt} 
    }
      \toprule
      & \multicolumn{3}{c@{\hskip 12pt}}{ \makecell[c]{\textbf{Fixed} \\\textbf{Domain Geometry}} } 
      & \multicolumn{2}{c@{\hskip 12pt}}{\makecell[c]{\textbf{Variable} \\\textbf{Domain Geometry}}} 
      & \multicolumn{3}{c@{\hskip 12pt}}{\makecell[c]{\textbf{Time }\\\textbf{Dependent}}} \\
      \cmidrule(lr){2-4} \cmidrule(lr){5-6} \cmidrule(lr){7-9}
      \textbf{MODEL}
        & \textbf{Poisson's}
        & \makecell[c]{\textbf{Darcy}-\textbf{Star}}
        & \textbf{Plate}
        & \makecell[c]{\textbf{Darcy-}\\\textbf{Pentagon}}
        & \makecell[c]{\textbf{Plate}\\\textbf{Variable}}
        & \makecell[c]{\textbf{Burgers}\\\textbf{1D}}
        & \makecell[c]{\textbf{Allen}-\\\textbf{Cahn}}
        & \makecell[c]{\textbf{Kuramoto}-\\\textbf{Sivashinsky}} \\
      \midrule
      PI-DCON                           & 0.6478  & 0.0411 & 0.0493 & 0.1360 & 0.1673 & - & –      &  - \\
      PI-GANO                           & –       & 0.0865 & 0.0224 & 0.1500 & 0.1324 & - & –      &  - \\
      PI-DeepONet                       & 0.6246  & 0.0947 & 0.0489 & 0.6172 & 0.4537 & - & –      &  - \\
      $\pi$G-Sp$^2$GNO-Encoder (ours)   & –       & 0.0369 & 0.0186 & \textbf{0.1014} & 0.1077 & - & –      &  - \\
      $\pi$G-Sp$^2$GNO (ours)           & \textbf{0.0162} & \textbf{0.0260} & \textbf{0.0159} & 0.2103 & \textbf{0.0890} & \textbf{0.3040} & \textbf{0.0685} &  \textbf{1.6128} \\
      \bottomrule
    \end{tabular}
  \end{adjustbox}
\end{table}

  
\subsubsection{Darcy flow problem on a star-shaped domain}
\label{subsec:darcy_star}
As the second example in this set, we consider the well-known Darcy flow problem on a star-shaped domain  \cite{PI-DCON}. 
The governing equation for this case is also an elliptic PDE of the form,
\begin{equation}
\label{eq:darcy_star}
    -\nabla \left( a( x, y) \nabla u( x, y)\right) = f( x, y), \quad (x, y) \in \mathcal{\mathcal{D}}
\end{equation}
where $a( x, y) = 1 {\text{ unit}}$ represents input permeability coefficients, $u( x, y)$ is the output pressure field, and $f( x, y) = 10 \text{ units}$ is the source term. The star-shaped domain $\mathcal{D}$ has a circular hole at the middle. The inner boundary of the hole is denoted as $\partial \mathcal{D}_1$ and the outer boundary of the star-shaped domain is denoted as $\partial \mathcal{D}_2$, such that the overall boundary is given by $\partial \mathcal{D} = \partial \mathcal{D}_1 \cup \partial \mathcal{D}_2$. 
A fixed zero boundary condition is applied on $\partial \mathcal{\mathcal{D}}_1$, while non-zero boundary  $g(x, y)$ is prescribed on $\partial \mathcal{\mathcal{D}}_2$, where  $g(x,y)$ is sampled from an one dimensional zero mean Gaussian Process with squared exponential kernel
\begin{equation}
    K(\xi, \xi') = \exp\left(-\frac{(\xi - \xi')^2}{2l^2}\right)
\end{equation}
With this setup, the objective here is to learn the solution operator that maps from the random boundary condition to the full-field solution by exploiting the proposed approach and the governing physics in Eq. \eqref{eq:darcy_star}.

\begin{figure}[ht]
  \centering
    \includegraphics[width=1.0\textwidth]{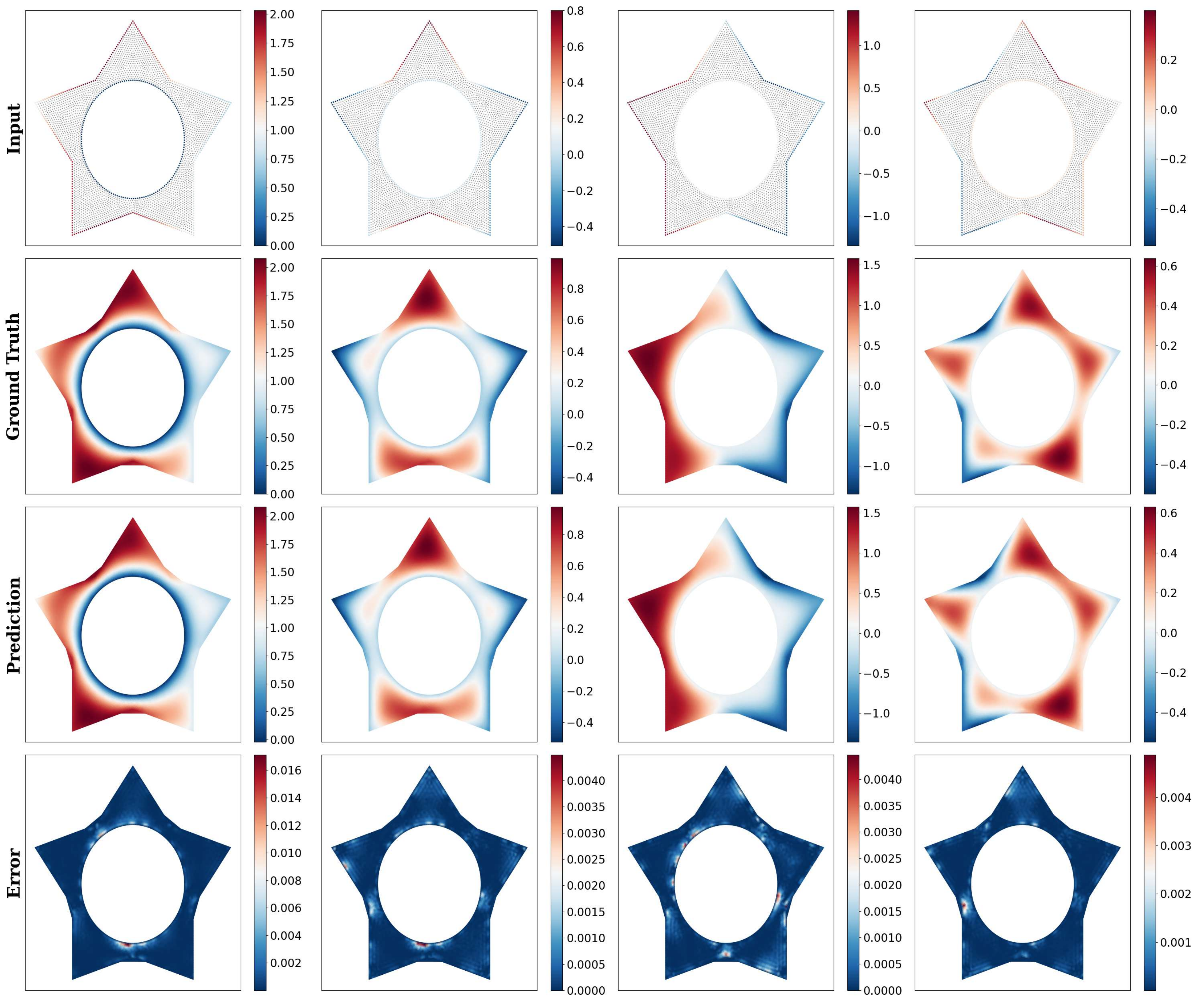}
  \caption{\textbf{Predictions for Darcy Equation on star-shaped domain}: The input BCs, ground truth, predictions, and errors of four different test examples are plotted in four columns. The contour plots in the first row represent the four examples' input boundary conditions. The respective ground truth and predictions for the same are plotted in the 2nd and 3rd rows, respectively. The corresponding point-wise errors are plotted in the fourth row.}
  \label{fig:darcy_star}
\end{figure}

Fig. \ref{fig:darcy_star} illustrates the solution field obtained using the proposed approach corresponding to four randomly sampled boundary conditions. We observe that the solution field obtained using the 
proposed $\pi$G-Sp$^2$GNO matches almost exactly with the ground truth results generated using Finite Element. Similar to previous example, we also compare the performance of the proposed approach as opposed to other benchmark models from the literature. The results of the comparative study are presented in Table \ref{table:performance_comparison}. 
Among all the models considered, the proposed $\pi$G-Sp$^2$GNO achieves the highest accuracy, surpassing its closest competitor, PI-GANO, by a margin of 58\%. This highlights the superior generalization capability and robustness of the proposed method across varying boundary conditions.

\subsubsection{2D Elastic Plate Equations}
\label{subsec:plate_dcon}
As the third benchmark problem in example set I, we consider the deformation of a two-dimensional elastic plate containing a circular hole at its center \cite{PI-DCON}. This classic elasticity problem serves to validate the capability of the proposed model in capturing stress concentrations arising from geometric discontinuities.
The governing equations for the in-plane deformation of the plate are derived from the equilibrium equations in linear elasticity under plane stress conditions. The coupled partial differential equations governing the displacement fields $u(x,y)$ and $v(x,y)$, representing displacements in the 
$x-$ and $y-$directions, respectively, are given by:
\begin{equation}\label{eq:eg3_ps}
\begin{aligned}
    \frac{E}{1 - \mu^2} \bigg( \frac{\partial^2 u( x, y)}{\partial  x^2} 
    + \frac{1 - \mu}{2} \frac{\partial^2 u( x,  y)}{\partial  y^2} 
    + \frac{1 + \mu}{2} \frac{\partial^2 v(x, y)}{\partial x \partial y} \bigg) &= 0, \quad (x, y) \in \mathcal{D}, \\[10pt]
    \frac{E}{1 - \mu^2} \bigg( \frac{\partial^2 v(x, y)}{\partial y^2} 
    + \frac{1 - \mu}{2} \frac{\partial^2 v(x, y)}{\partial x^2} 
    + \frac{1 + \mu}{2} \frac{\partial^2 u(x, y)}{\partial x \partial y} \bigg) &= 0, \quad (x, y) \in \mathcal{D},
\end{aligned}
\end{equation}
where $E$ denotes the Young's modulus, and $\mu$ is the Poisson's ratio. The computational domain $\mathcal{D}$ is a square plate of dimensions $20\; \text{mm} \times 20\; \text{mm}$, featuring a centrally located circular hole with a diameter of $5$ mm. A fixed (zero) displacement condition is imposed on the inner circular boundary $\partial D_H$, simulating a clamped hole,
\begin{equation}
\label{eq:zero_boundary_plate_dcon}
    u( x, y) = 0, \quad v( x,  y) = 0, \quad ( x, ,  y) \in \partial \mathcal{D}_H.
\end{equation}
On the left and right edges of the plate, denoted by $\partial\mathcal{D}_L$ and $\partial\mathcal{D}_R$, displacement boundary conditions are prescribed in terms of known displacement fields, $\left(u_L, v_L\right)$ and $\left(u_R, v_R\right)$, respectively: 
\begin{equation}
    \begin{aligned}
         u( x, y) &= u_L( x, y), \quad v( x, y) = v_L(x, y), \quad &&( x, y) \in \partial \mathcal{D}_L, \\
    u( x, y) &= u_R( x, y), \quad v( x, y) = v_R( x, y), \quad &&( x, y) \in \partial \mathcal D_R.
    \end{aligned}
\end{equation}
The top and bottom edges of the plate, corresponding to $\partial D_T$ and $\partial D_B$ are considered to be traction-free, which translates into the following constraints on the displacement gradients:
\begin{equation}
\label{eq:free_BC_dcon}
    \frac{\partial v( x,  y)}{\partial y} = 0, \quad 
    \frac{1}{2} \left[ \frac{\partial u( x, y)}{\partial y} + \frac{\partial v( x, y)}{\partial x} \right] = 0, \quad ( x,  y) \in \partial \mathcal{D}_{T} \cup \partial \mathcal{D}_{B}.
\end{equation}
The boundary conditions $u_L, v_L, v_R, v_R$ are considered to be random sampled from a GP with its covariance kernel dependent only on the $y$ coordinates,
\begin{equation}
\begin{aligned}
\label{eq:1d_gp}
    u_L, v_L, u_R, v_R \sim \mathcal{GP}(1, K( y, y')), \\[10pt]
    K( y, y') = \exp\left(-\frac{( y -  y')^2}{2l^2}\right).
\end{aligned}
\end{equation}
The objective is to learn the solution operator
$\mathcal{F}_{\bm \theta} : (u_L, v_L, u_R, v_R) \mapsto (u, v) $ by exploiting the proposed framework and the governing physics described in Eq. \eqref{eq:eg3_ps}. 

\begin{figure}[ht]
  \centering
  \includegraphics[width=1.0\textwidth]{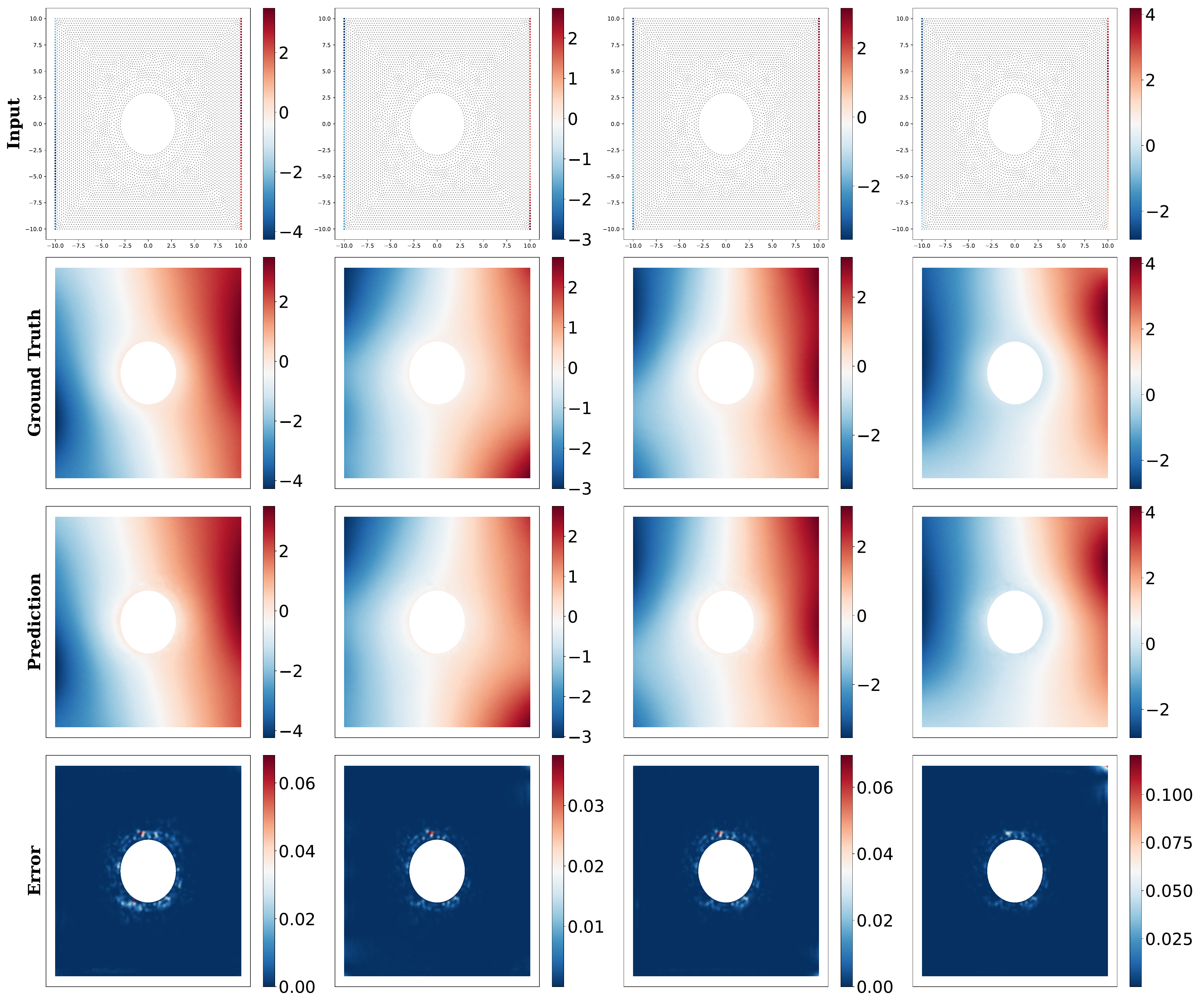}
  \caption{\textbf{Predictions of displacement component $\mathbf{u}$ for Elastic Plate Equations}: The input BCs, ground truth, predictions, and errors of four different test examples are plotted in four columns. The contour plots in the first row represent the four examples' Three Types of Encoders for three different kind of Problems. The respective ground truth and predictions for the same are plotted in the 2nd and 3rd rows, respectively. The corresponding point-wise squared errors are plotted in the fourth row.}
  \label{fig:plate_u}
\end{figure}

\begin{figure}[ht]
  \centering
  \includegraphics[width=1.0\textwidth]{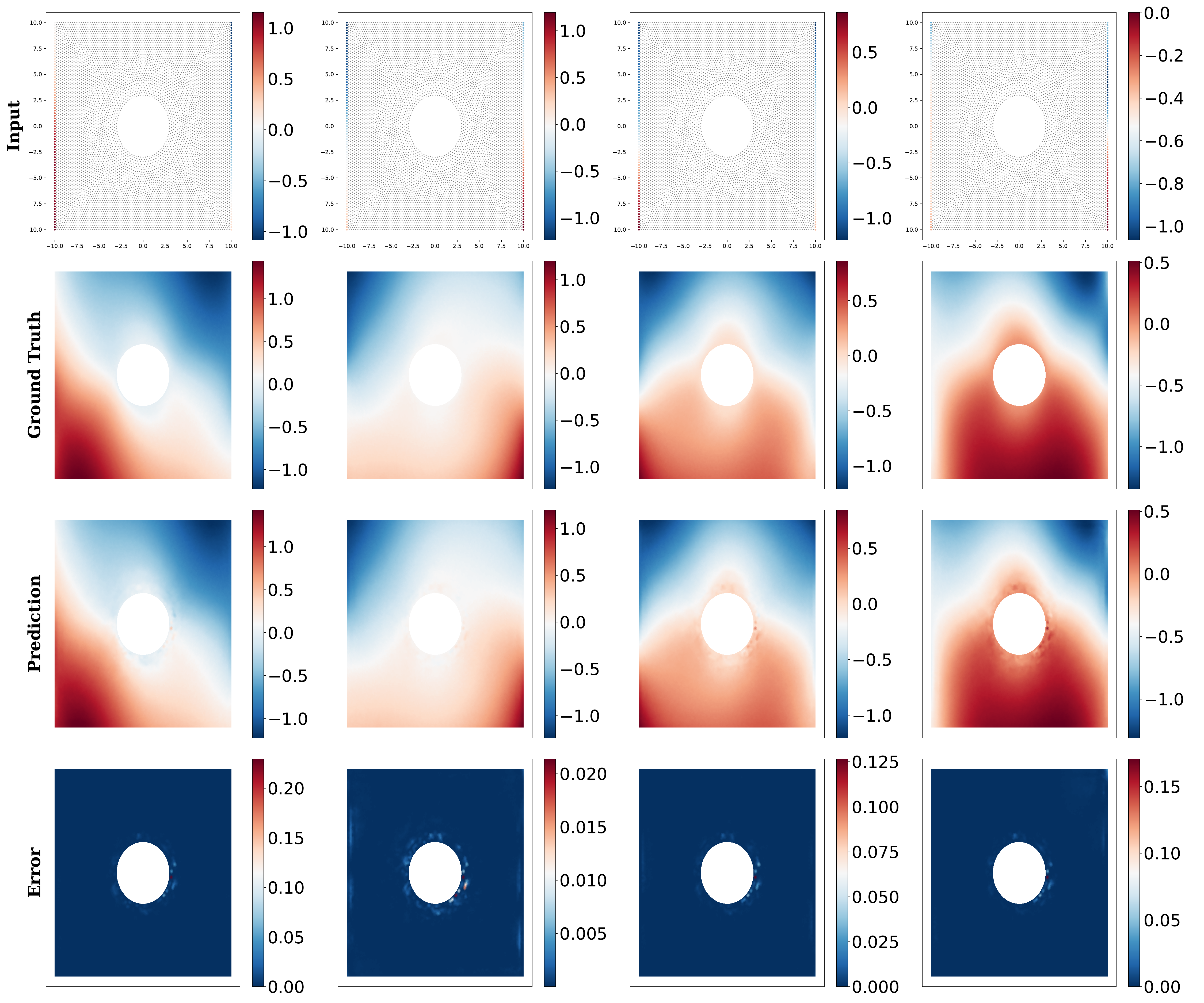}
  \caption{\textbf{Predictions of displacement component $\mathbf{v}$ for Elastic Plate Equations}: The input BCs, ground truth, predictions, and errors of four different test examples are plotted in four columns. The contour plots in the first row represent the four examples' input boundary conditions. The respective ground truth and predictions for the same are plotted in the 2nd and 3rd rows, respectively. The corresponding point-wise squared errors are plotted in the fourth row.}
  \label{fig:plate_v}
\end{figure}

Figs.~\ref{fig:plate_u} and ~\ref{fig:plate_v} illustrates the solution field obtained using the proposed approach corresponding to four randomly sampled boundary conditions. We observe that the results obtained using the proposed $\pi$G-Sp$^2$GNO (third row) matches well (visually) with the ground truth results (second row) obtained using finite element method. For quantitative assessment of the results obtained using the proposed approach are compared against state-of-the-art neural operators in Table \ref{table:performance_comparison}.  Similar to the previous two examples, the proposed approach yields the best result, outperforming its nearest competitor by 40.88\%, illustrating the excellent performance of the $\pi$G-Sp$^2$GNO for this problem defined using coupled differential equations.

\subsection{Example Set II: Time-independent problems with variation in geometry}
In the second set of examples, we consider time-independent problems characterized by variations in domain geometry. The primary objective in this category is to learn solution operators that generalize across different geometric configurations. This setting poses additional challenges due to the induced topological variability. We assess the performance of the proposed \(\pi\)G-Sp\(^2\)GNO in capturing such geometric generalization and benchmark it against existing state-of-the-art models.
\subsubsection{Darcy Equation on Variable Geometry}
As the first example in this set, we revisit the two-dimensional Darcy flow problem introduced earlier in Section~\ref{subsec:darcy_star}, now posed on a family of varying pentagonal domains \cite{PI-GANO}.
The geometries are procedurally generated through a two-step sampling strategy. First, a base pentagon is constructed by sampling five vertices uniformly from the circumference of a unit circle. Around each of these sampled vertices, we define a local perturbation region—a smaller circle of radius 0.25 centered at the vertex. To generate a specific geometry instance, five new vertices are uniformly sampled from these perturbation circles and connected via straight lines to form a pentagon. This process yields a diverse set of geometrical configurations, introducing nontrivial variability in the domain shape while preserving topological consistency.

To further increase the complexity of the learning task, the boundary conditions for each geometry are also varied. Specifically, the Dirichlet boundary conditions are modeled as realizations of a Gaussian Process with a squared exponential kernel, ensuring smooth but spatially diverse boundary profiles across the dataset.
Within this setup, our objective is to learn the parametric solution operator 
\begin{equation}
\label{eq:geo_dependent_sol_op}
    \mathcal{F}: \left(g(x, y), \mu \right) \mapsto u(x, y),
\end{equation}
where $g(x, y)$ denotes the spatially varying boundary condition, $\mu$ encodes the geometry of the domain (parameterized by the sampled vertices), and $u(x, y)$ is the resulting solution field of the Darcy flow equation. 
\begin{figure}[ht]
  \centering
  \includegraphics[width=1.0\textwidth]{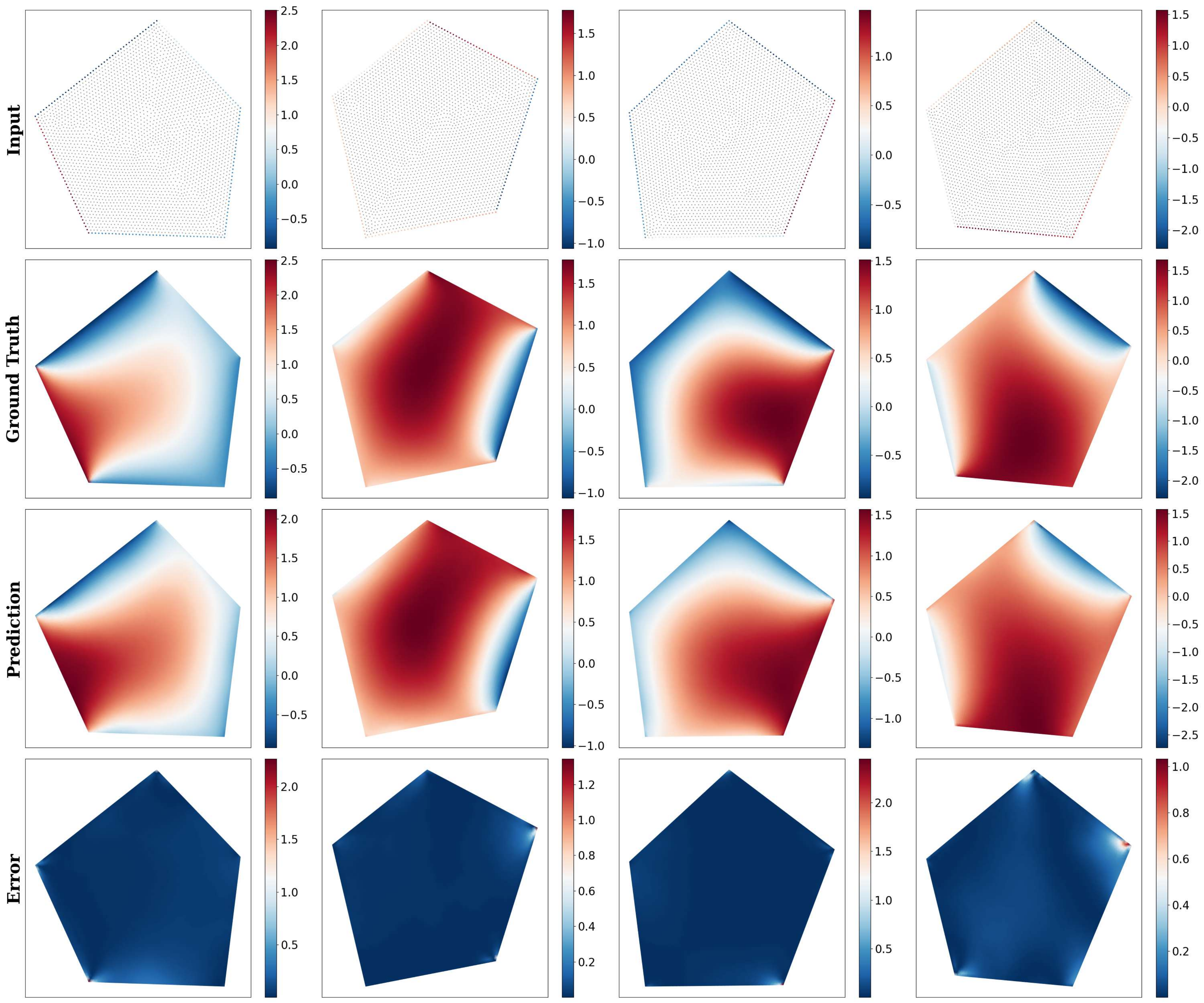}
  \caption{\textbf{Predictions for Darcy Equation on Variable Domain Geometries}: The input BCs, ground truth, predictions, and errors of four different test examples are plotted in four columns. The contour plots in the first row represent the four examples' input boundary conditions. The respective ground truth and predictions for the same are plotted in the 2nd and 3rd rows, respectively. The corresponding point-wise squared errors are plotted in the fourth row.}
  \label{fig:darcy_pentagon}
\end{figure}
Figure~\ref{fig:darcy_pentagon} illustrates the predictive performance of the proposed $\pi$G-Sp$^2$GNO framework on the Darcy flow problem defined over varying pentagonal domains. The figure presents results for four representative test samples, each characterized by distinct geometries and boundary conditions randomly generated as described above. The second row displays the ground truth solutions obtained using the finite element method, while the third row shows the corresponding predictions produced by the proposed $\pi$G-Sp$^2$GNO model. A visual comparison reveals excellent agreement between the predicted and reference solution fields. This close correspondence is further corroborated by the error contours (bottom row), which show that the pointwise error remains consistently low across all test instances.
To quantitatively evaluate the performance, we compare the N-MSE of $\pi$G-Sp$^2$GNO against state-of-the-art baseline models from the literature. The proposed approach achieves the lowest N-MSE of 0.1227, outperforming PI-DCON (N-MSE = 0.1360) and PI-GANO (N-MSE = 0.1500). Furthermore, we observe that in contrast to the earlier examples, the variant of $\pi$G-Sp$^2$GNO that incorporates an encoder yields superior accuracy in this case. This highlights the importance of the encoder in effectively capturing the impact of geometric variations on the solution operator.

\subsubsection{Plate Problem on Variable Geometry}
To consolidate our claim about the ability of proposed $\pi$G-Sp$^2$GNO in handling variation in geometry, we revisit the plate problem in \ref{subsec:plate_dcon} posed on variable geometries \cite{PI-GANO}. We consider the domain to be a $20$ mm $\times$ $20$ mm plate with four holes of different radii at different locations. Each domain sample was generated by following a two-step procedure. First, we generate a base domain geometry consisting of four circular holes of radii $1.5$ mm located at four corners of a square of side length $10$ mm centered at $(0, 0)$. The location of these four circular holes are $(-5,5)$, $(5,5)$, $(5,-5)$, and $(-5, -5)$. In the second step, the center of each hole was sampled from the perimeter of these four circles of the base geometry, and their corresponding radii were sampled from the uniform distribution $\mathcal{U}(0.8, 1.5)$. Each sampled domain has four circular holes of varying radius at different locations of the plate. Boundary conditions of the problem was defined similarly to the plate problem described in subsection \ref{subsec:plate_dcon}, with the left and right edges of each domain geometry being sampled from a 1D Gaussian process described in Eq.~\eqref{eq:1d_gp}. 
\begin{figure}[ht]
  \centering
  \includegraphics[width=1.0\textwidth]{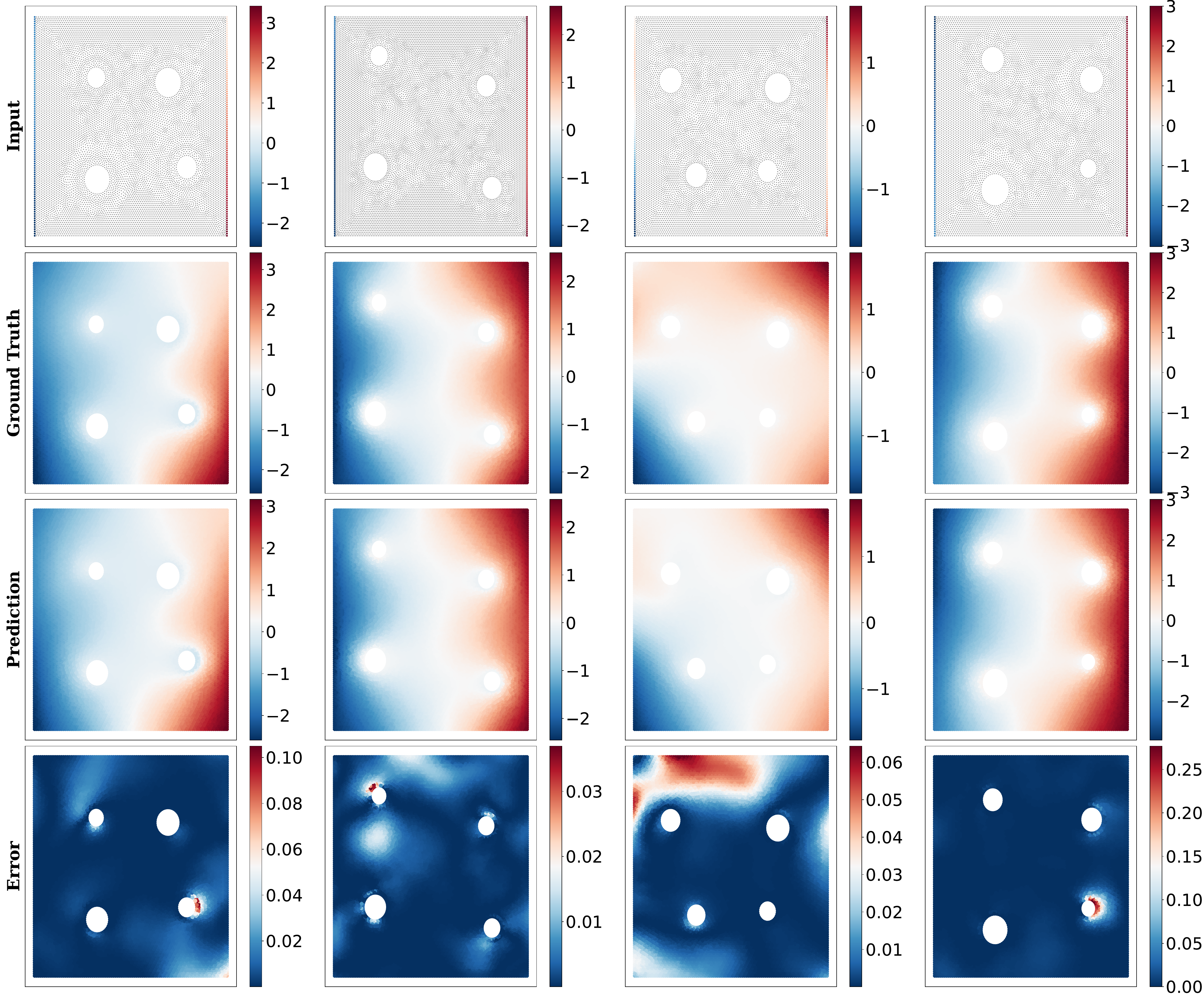}
  \caption{\textbf{Predictions of displacement component $\mathbf{u}$ Elastic Plate Equations with variable domain geometries}: The input BCs, ground truth, predictions, and errors of four different test examples are plotted in four columns. The contour plots in the first row represent the four examples' input boundary conditions. The respective ground truth and predictions for the same are plotted in the 2nd and 3rd rows, respectively. The corresponding point-wise squared errors are plotted in the fourth row.}
  \label{fig:plate_variable_u}
\end{figure}

\begin{figure}[ht]
  \centering
  \includegraphics[width=1.0\textwidth]{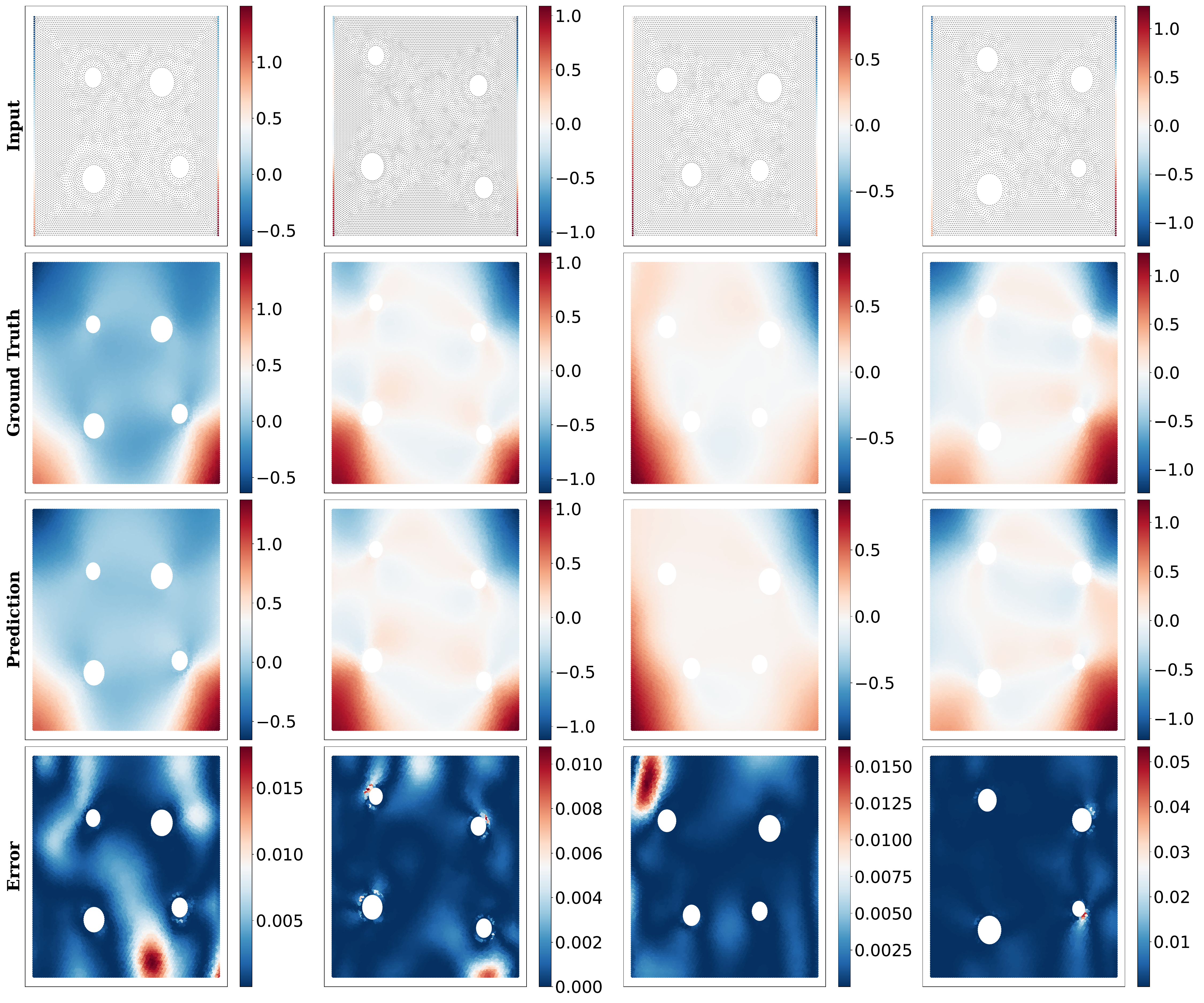}
  \caption{\textbf{Predictions of displacement component $\mathbf{v}$ for Elastic Plate Equations with variable domain geometries}: The input BCs, ground truth, predictions, and errors of four different test examples are plotted in four columns. The contour plots in the first row represent the four examples' input boundary conditions. The respective ground truth and predictions for the same are plotted in the 2nd and 3rd rows, respectively. The corresponding point-wise squared errors are plotted in the fourth row.}
  \label{fig:plate_variable_v}
\end{figure}
We train our model to learn the solution operator described in Eq.~\eqref{eq:geo_dependent_sol_op}. Figs. ~\ref{fig:plate_variable_u} and ~\ref{fig:plate_variable_v} illustrate the excellent capability of $\pi$G-Sp$^2$GNO in learning solution operators for geometry-dependent problems from governing physics.  Table~\ref{table:performance_comparison} 
provides a quantitative assessment of the proposed approach against other state-of-the-art neural operators. We observe that the proposed $\pi$G-Sp$^2$GNO achieves an N-MSE of $0.0890$, outperforming PI-DCON (N-MSE = $0.1673$) and PI-GANO (N-MSE = $0.1324$).


\subsection{Example set III: Time-dependent problems}
The third set of examples addresses time-dependent problems, where the objective is to learn the spatiotemporal evolution of the solution field given initial conditions. These problems are inherently challenging due to the added temporal complexity and the presence of long-range dependencies. We note that the 
state-of-the art algorithms considered in example sets I and II, cannot handle time-dependent problems, and hence, we focus on demonstrating the efficacy of the proposed \(\pi\)G-Sp\(^2\)GNO in capturing complex spatiotemporal dynamics.

\subsubsection{1D Burgers’ Equation}
As the first example in this set, we evaluate the performance of the proposed approach on the one-dimensional viscous Burgers’ equation \cite{ArDenseED},
\begin{align}
\frac{\partial u}{\partial t} + u \frac{\partial u}{\partial x} - \nu \frac{\partial^2 u}{\partial x^2} &= 0, \label{eq:burgers}\\
u(0, t) &= u(L, t), \quad x \in [0, L], \quad t \in [0, T]. \label{eq:bc}
\end{align}
Here, \( u(x, t) \) represents the velocity field, and \( \nu \) denotes the kinematic viscosity.

\begin{figure}[ht]
  \centering
  \includegraphics[width=1.0\textwidth]{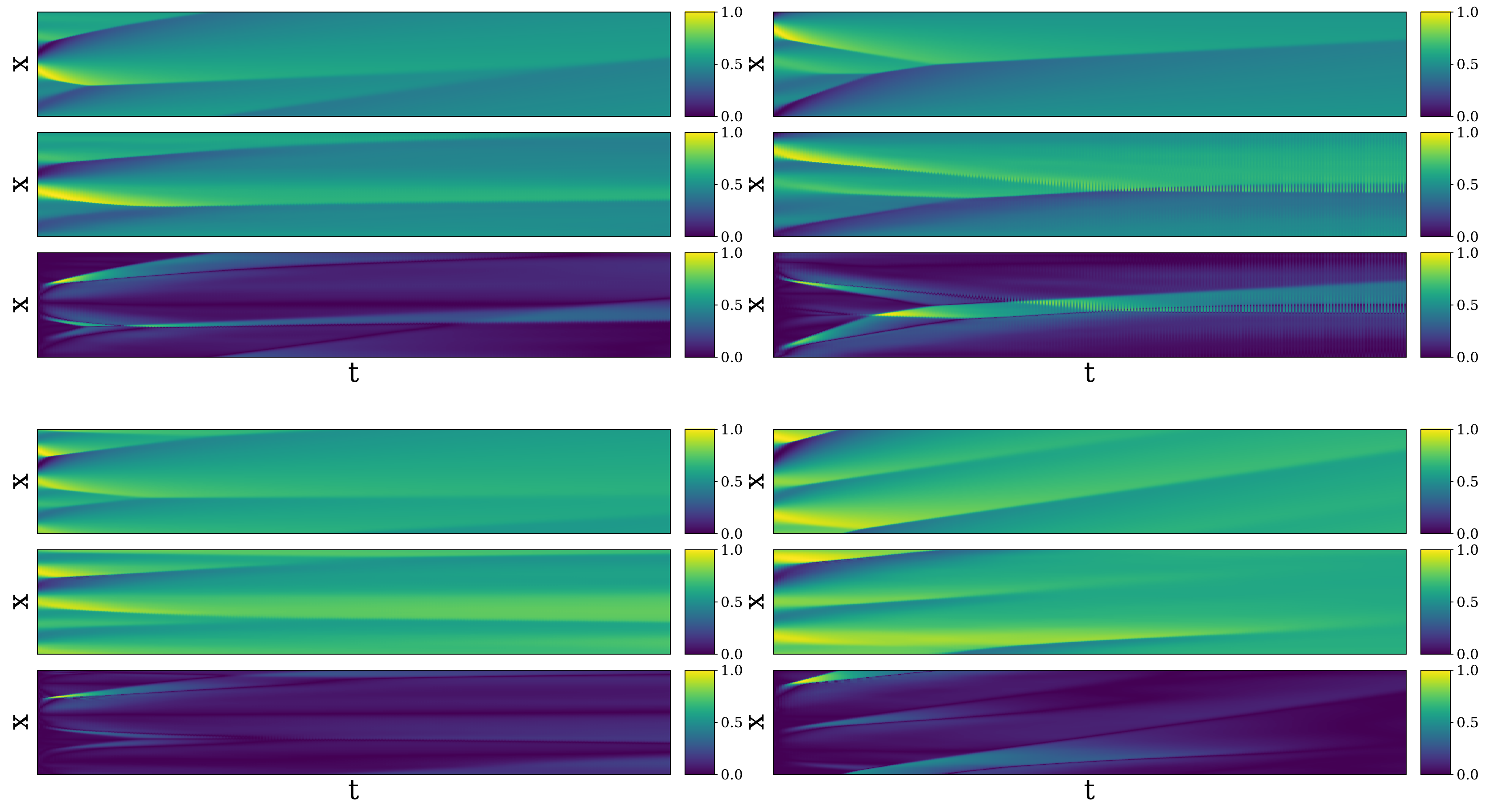}
  \caption{\textbf{Predictions for 1D Burgers Equations}: The ground truth, predictions, and errors of four different test examples are plotted in four different subplots. In each subplot, the contour plots in the first row represent ground truth. The predictions and corresponding point-wise squared errors for the same are plotted in the 2nd and 3rd rows, respectively.}
  \label{fig:burgers1d}
\end{figure}
We employ a highly variable initial condition constructed through a random Fourier series:
\begin{equation}
w(x) = a_0 + \sum_{l=1}^{L} a_l \sin(2 l \pi x) + b_l \cos(2 l \pi x),
\end{equation}
where the coefficients \( a_l \) and \( b_l \) are sampled from a standard normal distribution \( \mathcal{N}(0, 1) \). The initial velocity field is then normalized and shifted as follows:
\begin{equation}
u(x, 0) = \frac{2 w(x)}{\max_x |w(x)|} + c,
\end{equation}
with \( c \) drawn from a uniform distribution \( \mathcal{U}(-1, 1) \) to introduce additional randomness. In our experiments, we set \( L = 1 \) and the viscosity \( \nu = 0.0025 \), resulting in rich nonlinear dynamics characterized by the formation, interaction, and merging of multiple shocks over time. These challenging solution trajectories make the Burgers’ equation an ideal testbed for assessing the capability of $\pi$G-Sp$^2$GNO in learning intricate spatiotemporal patterns from the governing physics. 

As described in Section~\ref{subsec:problem_setup_for_time_depndent}, the model is trained to predict the system response at the next time step using its own predictions from the past \( k = 10 \) time steps as input. This auto-regressive strategy enables the model to generate long-term predictions, starting from the initial condition \( u_0 \) and progressing up to the desired time steps during training. Here, we progress until 200 time steps during training. Figure~\ref{fig:burgers1d} illustrates the ground truth, predicted trajectories, and corresponding error fields for four representative initial conditions. The results demonstrate that the proposed \(\pi\)G-Sp\(^2\)GNO effectively captures the spatiotemporal dynamics of the 1D Burgers equation for up to 400 time steps, with a temporal resolution of \( \Delta t = 0.005 \, \text{s} \). The N-MSE, reported in Table~\ref{table:performance_comparison}, further confirms the model’s high predictive accuracy in this challenging time-dependent setting.

\subsubsection{1D Kuramoto-Sivashinsky Equation}
As the second example in this set, we evaluate the performance of $\pi$G-Sp$^2$GNO on the one-dimensional Kuramoto–Sivashinsky (K–S) equation \cite{ArDenseED}. The K–S equation is well-known for exhibiting rich spatiotemporal dynamics, particularly chaotic behavior in large periodic domains. It serves as a canonical model for investigating weak turbulence and complex nonlinear phenomena, with applications spanning flame front propagation, plasma instabilities, and chemical reaction systems. The governing equation is expressed as:
\begin{align}
\frac{\partial u}{\partial t} + u \frac{\partial u}{\partial x} + \frac{\partial^2 u}{\partial x^2} + \nu \frac{\partial^4 u}{\partial x^4} &= 0, \label{eq:ks}\\
u(0, t) &= u(L, t), \quad x \in [0, L], \quad t \in [0, T]. \label{eq:ks_bc}
\end{align}
Here, \( u(x, t) \) represents the solution field and \( \nu \) is the hyper-viscosity coefficient, which we set to \( \nu = 1 \) throughout this study. The spatial domain is defined as \( x \in [0, 22\pi] \),  a range known to induce fully chaotic dynamics in the system. 
This domain is discretized into 96 uniformly spaced cells, with a constant time step of \( \Delta t = 0.1 \). To isolate the system's sustained chaotic behavior, we discard transient dynamics and consider only the solutions for \( t \geq 100 \).  The objective of this experiment is to assess the capability of $\pi$G-Sp$^2$GNO in accurately capturing the long-term chaotic evolution of the K–S system from complex initial conditions. 
\begin{figure}[ht]
  \centering
  \includegraphics[width=1.0\textwidth]{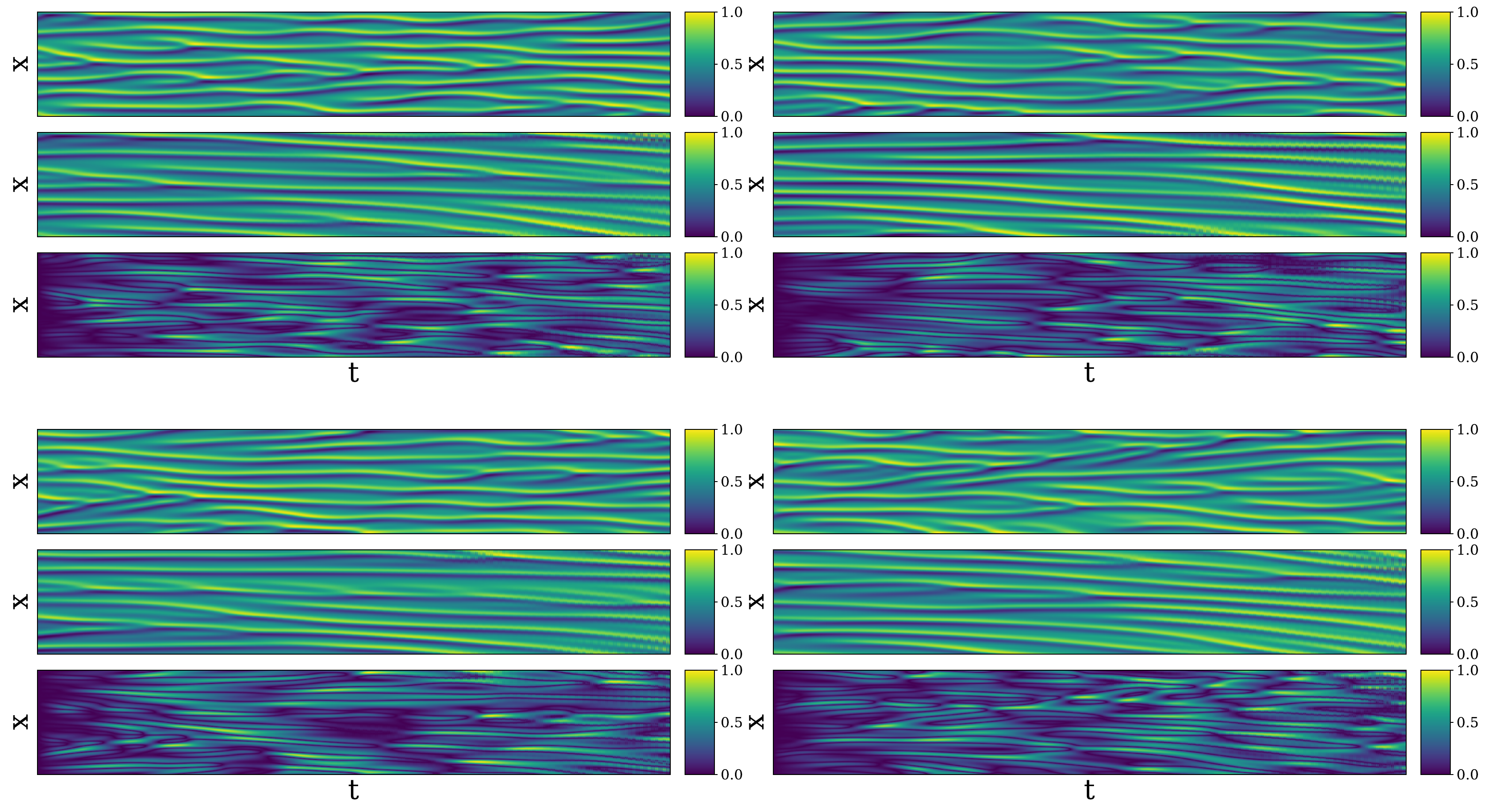}
  \caption{\textbf{Predictions for 1D Kuramoto-Sivashinsky Equations}: The ground truth, predictions, and errors of four different test examples are plotted in four different subplots. In each subplot, the contour plots in the first row represent ground truth. The predictions and corresponding point-wise squared errors for the same are plotted in the 2nd and 3rd rows, respectively.}
  \label{fig:KS}
\end{figure}
The model is trained to predict the system response at the next time step, using its own predictions from the previous $k=10$ time steps as input. This autoregressive framework enables the model to forecast the system evolution over long time-horizon. The contour plots in Fig.~\ref{fig:KS} and the normalized mean-squared error (N-MSE) values presented in Table~\ref{table:performance_comparison} demonstrate the strong predictive capability of the proposed model.

\subsubsection{Time dependent Allen Cahn's Equation}
As the final example in this set, we consider the Allen–Cahn equation defined on a regular two-dimensional domain \cite{PIWNO},
\begin{equation}\label{eq:AC}
\begin{aligned}
\frac{\partial u(x, y, t)}{\partial t} &= \epsilon \Delta u(x, y, t) + u(x, y, t) - u(x, y, t)^3, && x, y \in [0, 1], \\
u(x, y, 0) &= u_0(x, y), && x, y \in [0, 1],
\end{aligned}
\end{equation}
where \( \epsilon \) is a positive real number representing the viscosity coefficient, and \( u(x, y, t) \) denotes the state of the system at spatial location \( (x, y) \) and time \( t \).
The initial condition \( u_0(x, y) \) is modeled as a realization of a Gaussian process defined by a two-dimensional anisotropic Matérn random field with the Matérn covariance kernel:
\begin{equation}
\rho(h) = \sigma^2 \cdot \frac{2^{1-\nu}}{\Gamma(\nu)} \left( \frac{\sqrt{2\nu} h}{\ell} \right)^{\nu} K_{\nu} \left( \frac{\sqrt{2\nu} h}{\ell} \right),
\end{equation}
where \( h \) denotes the anisotropically scaled distance, \( \nu = 3 \) is the smoothness parameter, \( \ell = (\lambda_x, \lambda_y) = (0.5, 0.5) \) defines the correlation length scales in the \( x \)- and \( y \)-directions, \( \sigma^2 = 0.01 \) is the variance, and \( K_{\nu} \) is the modified Bessel function of the second kind.
The goal of this experiment is to evaluate the ability of the proposed $\pi$G-Sp$^2$GNO framework to learn the time-dependent solution operator for the Allen–Cahn system.

\begin{figure}[ht]
  \centering
  \includegraphics[width=\textwidth]{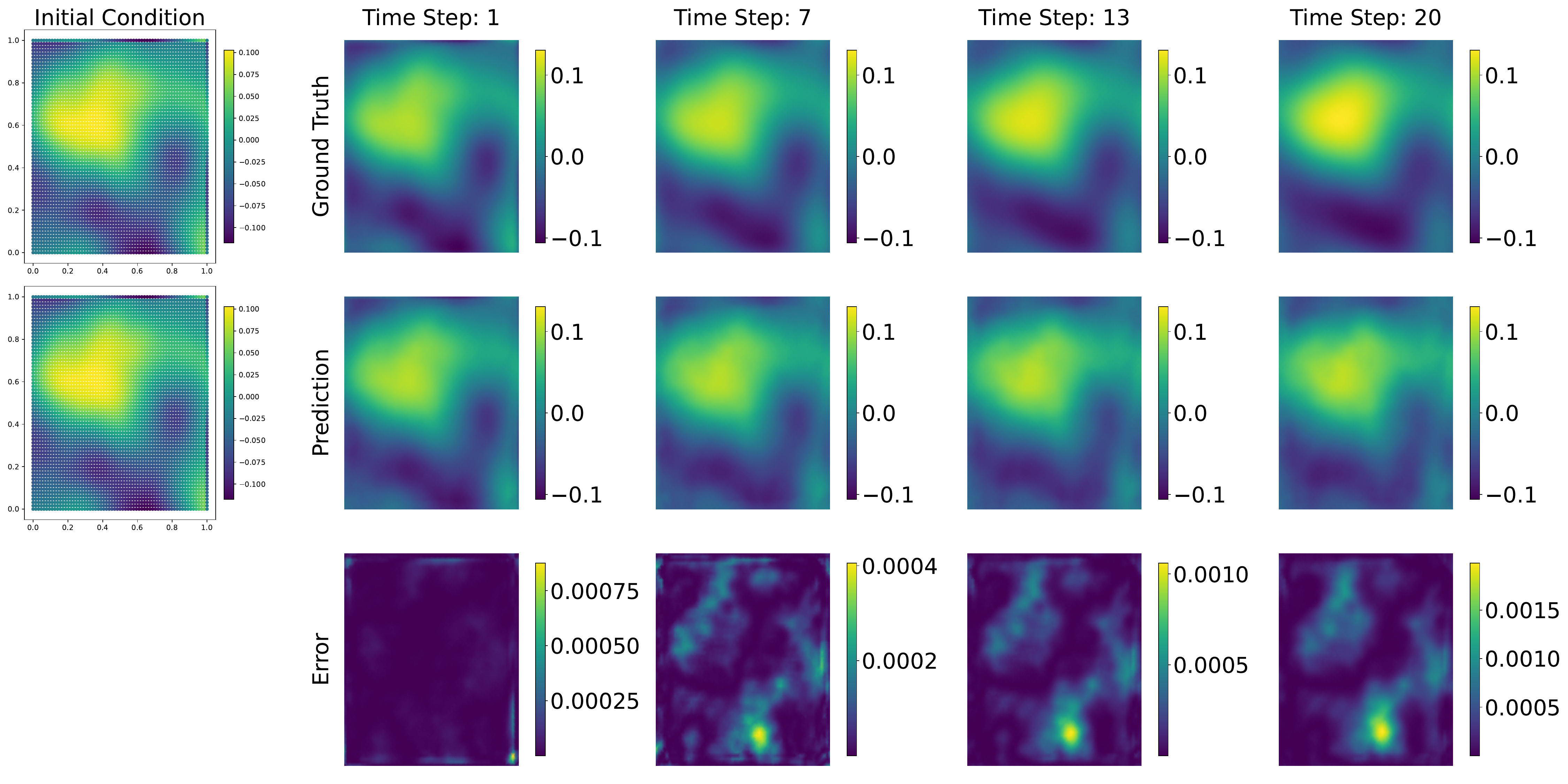}
  \caption{\textbf{Predictions for the Allen-Cahn Problem only using the initial state $u_0$ as input}: From left to right, the plots in the first column are the initial conditions. Each of the next ten columns corresponds to time-steps 1 to 19. The first row showcases the ground truth at each time step. The second and third rows show the prediction and point-wise squared errors for each time step.}
  \label{fig:allen_cahn}
\end{figure}

\begin{figure}[ht]
  \centering
  \includegraphics[width=\textwidth]{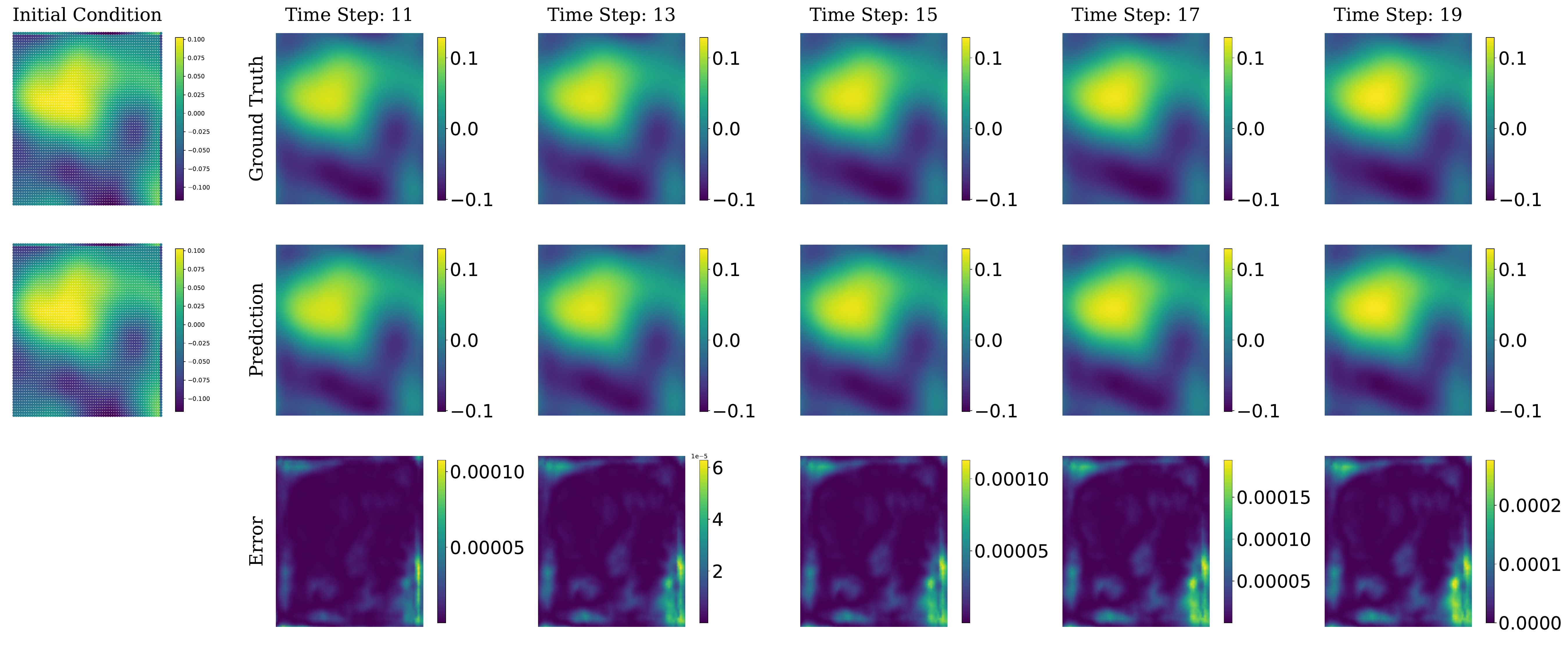}
  \caption{\textbf{Predictions for the Allen-Cahn Problem using the response of first 10 time-steps $[u_0,u_1,\cdots, u_{10}]$ as input}: From left to right, the plots in the first column are the initial conditions. Each of the next ten columns corresponds to time-steps 1 to 19. The first row showcases the ground truth at each time step. The second and third rows show the prediction and point-wise squared errors for each time step.}
  \label{fig:allen_cahn_data_first_10}
\end{figure}

We consider two distinct scenarios to evaluate the performance of the proposed $\pi$G-Sp$^2$GNO framework. In the first scenario, the model is trained using only the initial condition \( u_0 \) and the governing physics described in Eq.~\eqref{eq:AC}, following the procedure outlined in Section~\ref{subsec:problem_setup_for_time_depndent}. 
In the second scenario, we generate the system response for the first \( k = 10 \) time steps using a spectral Galerkin method with semi-implicit time integration. The model is then trained to learn the time-dependent solution operator for the remaining \( N-k \) time steps in a recursive manner, using the same physics-informed loss function as in the first setup.
Fig.~\ref{fig:allen_cahn} illustrates the performance of the proposed method in capturing the temporal evolution of the solution field for the Allen–Cahn problem. The results predicted by $\pi$G-Sp$^2$GNO (second row) exhibit close agreement with the ground truth obtained using the spectral Galerkin method (first row), as further supported by the error contours shown in the third row of Fig.~\ref{fig:allen_cahn}.
Additionally, Fig.~\ref{fig:allen_cahn_data_first_10} presents the results for the second setup, where the responses from the first 10 time steps are provided to the model. Quantitatively, the proposed approach achieves a normalized mean-squared error (N-MSE) of 0.0685 when trained using only the initial condition, and an N-MSE of 0.088 when the responses from the first 10 time steps are used as input. Overall, these results demonstrate the effectiveness of the proposed $\pi$G-Sp$^2$GNO framework in learning time-dependent solution operators

\section{Conclusion}\label{sec:concl}

We introduced $\pi$G-Sp$^2$GNO, a physics- and geometry-aware spatio-spectral graph neural operator for learning solution operators of both time-independent and time-dependent partial differential equations across regular, irregular, and variable geometries. The model incorporates physical laws directly into the training loss, eliminating the need for supervised data and enabling learning solely from the governing equations. This makes $\pi$G-Sp$^2$GNO particularly well-suited for scientific domains where labeled data is scarce or expensive to obtain.
To compute spatial gradients without relying on automatic differentiation, we employ a stochastic projection method, making the approach more scalable and adaptable to a wide range of PDE formulations. Numerical experiments show that $\pi$G-Sp$^2$GNO successfully handles a diverse set of problems including chaotic systems, variable geometries, and long time horizons, achieving high accuracy while maintaining complete data efficiency.

Despite these advantages, the approach has limitations. Its performance is sensitive to the choice of certain hyperparameters, such as the number of nearest neighbors used for graph construction and the weights assigned to various components of the physics-based loss. Tuning these parameters can be computationally expensive and may require significant trial and error. Future work will focus on enabling end-to-end learning of these parameters to reduce manual intervention and improve overall robustness.
In summary, $\pi$G-Sp$^2$GNO offers a unified, data-efficient, and physics-consistent framework for solving complex PDEs on arbitrary domains, making it a valuable tool in the development of next-generation scientific machine learning models.

\section*{Acknowledgements}
SC acknowledges the financial support received from the Ministry of Ports and Shipping via letter number ST-14011/74/MT (356529) and the Anusandhan National Research Board via grant number CRG/2023/007667.

\section*{Code availability}
Upon acceptance, all the source codes to reproduce the results in this study will be made available on request.

\section*{Competing interests} 
The authors declare no competing interests.


\end{document}